\documentclass[10pt,twocolumn,letterpaper]{article}

\usepackage{iccv}              

\usepackage{amssymb}
\usepackage{pifont}
\newcommand{\cmark}{\ding{51}}%
\newcommand{\xmark}{\ding{55}}%
\usepackage{makecell}
\usepackage{multirow}

\newcommand{\datasetname}{\textsc{Museum-65}\xspace}
\usepackage{enumitem}

\definecolor{cvprblue}{rgb}{0.21,0.49,0.74}
\usepackage[pagebackref,breaklinks,colorlinks,allcolors=cvprblue]{hyperref}
\usepackage{epigraph} 
\usepackage{amssymb}
\usepackage{fontawesome5}

\def\paperID{11539} 
\def\confName{ICCV}
\def\confYear{2025}

\title{Understanding Museum Exhibits using Vision-Language Reasoning}

\author{%
  Ada-Astrid Balauca$^{1,}$\thanks{Equal Contribution \qquad \faIcon[regular]{envelope}\, \texttt{astrid.mokanu@gmail.com}}\textsuperscript{\textasteriskcentered$,$ \faIcon[regular]{envelope}} \quad
  Sanjana Garai$^{1,3,\text{\textasteriskcentered}}$ \quad
  Stefan Balauca$^{1}$ \quad
  Rasesh Udayakumar Shetty$^{3}$ \vspace{0.15em}\\
  Naitik Agrawal$^{3}$ \quad
  Dhwanil Subhashbhai Shah$^{3}$ \quad
  Yuqian Fu$^{1}$ \quad
  Xi Wang$^{1,2}$ \vspace{0.15em}\\
  Kristina Toutanova$^{1,4}$ \quad
  Danda Pani Paudel$^{1}$ \quad
  Luc Van Gool$^{1}$ \vspace{0.65em}\\
  $^{1}$INSAIT, Sofia University ``St. Kliment Ohridski'', Bulgaria \hspace{1em}
  $^{2}$ETH Zürich, Switzerland \hspace{1.5em}\\
  $^{3}$Indian Institute of Technology, Varanasi (IIT BHU) \hspace{1em}
  $^{4}$Google DeepMind \hfil
}

\AddToHook{cmd/appendix/before}{%
    \crefalias{section}{appendix}%
    \crefalias{subsection}{appendix}
}

\begin{document}
\crefname{appendix}{App.}{Apps.}
\Crefname{appendix}{App.}{Apps.}
\maketitle
\begin{abstract}
Museums serve as repositories of cultural heritage and historical artifacts from diverse epochs, civilizations, and regions, preserving well-documented collections that encapsulate vast knowledge, which, when systematically structured into large-scale datasets, can train specialized models. Visitors engage with exhibits through curiosity and questions, making expert domain-specific models essential for interactive query resolution and gaining historical insights. Understanding exhibits from images requires analyzing visual features and linking them to historical knowledge to derive meaningful correlations. We facilitate such reasoning by (a) collecting and curating a large-scale dataset of 65M images and 200M question-answer pairs for exhibits from all around the world; (b) training large vision-language models (VLMs) on the collected dataset; (c) benchmarking their ability on five visual question answering tasks, specifically designed to reflect real-world inquiries and challenges observed in museum settings.
The complete dataset is labeled by museum experts, ensuring the quality and the practical significance of the labels. 
We train two VLMs from different categories: BLIP~\cite{li2022blip} with vision-language aligned embeddings, but lacking the expressive power of large language models, and the LLaVA~\cite{liu2023visualinstructiontuning} model, a powerful instruction-tuned LLM enriched with vision-language reasoning capabilities.  
Through extensive experiments, we find that while both model types effectively answer visually grounded questions, large vision-language models excel in queries requiring deeper historical context and reasoning. We further demonstrate the necessity of fine-tuning models on large-scale domain-specific datasets by showing that our fine-tuned models significantly outperform current SOTA VLMs in answering questions related to specific attributes, highlighting their limitations in handling complex, nuanced queries. Our dataset, benchmarks, and source code 
are available at: \href{https://github.com/insait-institute/Museum-65}{insait-institute/Museum-65}.  
\end{abstract}
\vspace{-2.5 mm}
\vspace{-5mm}
\section{Introduction}
\vspace{-1 ex}
\label{sec:intro}
\begin{figure*}
    \centering
    \includegraphics[clip, trim=0cm 0cm 0cm 0.5cm, width=0.94\textwidth]{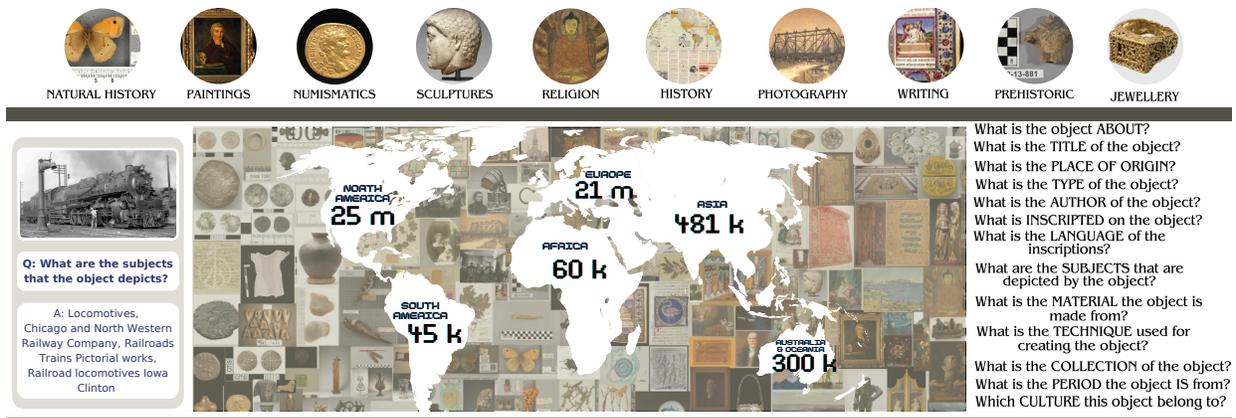}
    \vspace{-3mm}
    \caption{\textbf{Dataset composition.} \datasetname covers a wide range of exhibit categories (top), e.g arts, historical/pre-historical, natural sciences, and contains a \textbf{large number of images from around the globe}. Each image is paired with multiple questions exploring subjects like Title, Creator, Period, Techniques, Culture, Inscriptions, etc. (right). A sample image with a question and answer is shown on the left. }
    \label{fig:teaser}
    \vspace{-6mm}
\end{figure*}
We release a high-quality, large-scale dataset and demonstrate through experiments that training large VLMs on it enables museum artifact understanding, for visually understanding exhibit images through visual question answering. VLMs like CLIP~\cite{radford2021learning}, Gemini~\cite{team2023gemini}, and LLaVA~\cite{liu2023visualinstructiontuning} have demonstrated strong capabilities in learning from large-scale noisy image-text data, improving visual understanding through natural language and bridging the gap between textual annotations and images~\cite{wei2023improving,wang2022medclip,cui2024learning,zhai2022lit,pham2023combined,rasheed2023fine,meng2023foundation,lin2024fine,ramesh2021zero}. However, these models~\cite{li2022blip,liu2023visualinstructiontuning} struggle in domains like museums, which require detailed, interdisciplinary knowledge and structured attribute prediction, such as age, origin, material, and cultural relevance~\cite{becattini2023viscounth,marty2008museum,nishanbaev2019survey}. While pre-trained VLMs are effective for tasks like object detection~\cite{gu2021open,bangalath2022bridging,zhou2022detecting} and semantic segmentation~\cite{ding2021decoupling,Boyi2022,zhou2022extract}, more complex multi-modal tasks demand advanced reasoning across visual and textual domains~\cite{ju2022prompting,ni2022expanding,wang2021actionclip,rasheed2023fine}.  
\\
Visual Question Answering (VQA) is a key multi-modal task  explored in works like~\cite{antol2015vqa,singh2019towards,zhu2016visual7w,bigham2010vizwiz,malinowski2015ask,sadeghi2015viske,becattini2023viscounth}. In the cultural heritage domain, VQA can enhance museum engagement, but a large-scale dataset covering diverse artifacts with both visual and textual data is lacking. Existing datasets primarily focus on art~\cite{ruta2022stylebabel,52wilber2017bam,strezoski2018omniart} and are often used for image generation and style transfer~\cite{gatys2016image,kwon2022clipstyler,deng2022stytr2,ramesh2021zero}, failing to capture deeper exhibit-context relationships.  
\\
In this work, we collect a novel large-scale \textbf{multilingual dataset} \datasetname with high-quality images and extensive textual information for a wide range of museum artifacts, totaling 65M images and 200M question-answer pairs. We curate and use it to fine-tune VLMs, BLIP and LLaVA, to enable a better understanding of museum exhibits. The textual information of \datasetname reflects the viewpoint of knowledgeable museum experts, providing both depth and breadth for effective AI training. We further design 5 real-world tasks: general VQA, category-wise VQA, MultiAngle -- questions using images from different viewpoints, Visually Unanswerable Questions -- complex questions requiring the use of general knowledge, and MultiLanguage -- questions in languages other than the English. We also conduct an ablation study on place of origin to assess potential regional biases. We also show in ~\cref{tab:model_comparison} that the existing vision-language models  perform poorly, using questions for place of origin and object title.
\begin{table}[h]
    \centering
    \vspace{-3mm}
    \resizebox{0.8\linewidth}{!}{
    \begin{tabular}{l c cc}
        \toprule
        \multirow{2.4}{*}{Model Name} & \multirow{2.4}{*}{Zero-Shot} & \multicolumn{2}{c}{Attribute} \\
        \cmidrule(lr){3-4}
        &  & Title & Place \\
        \midrule
        GPT-4o & \cmark & 22.03 & 33.33 \\
        Claude-3-7-sonnet & \cmark & 21.89 & 40.43 \\
        Llama-3.2-90b-vision & \cmark & 16.84 & 29.58  \\
        Gemini 1.5B flash & \cmark & 27.08 & 32.98 \\
        \midrule
        LLaVa nofinetune & \cmark & 10.13 & 23.42 \\
        LLaVa-ours (20mn 1ep) & \xmark & \textbf{57.00} & \textbf{70.00} \\
        BLIP nofinetune & \cmark & 3.00 & 5.00 \\
        BLIP-ours (20mn 5ep) & \xmark & \textbf{52.00} & \textbf{61.00} \\
        \bottomrule
    \end{tabular}}
    \vspace{-2 ex}
    \caption{\textbf{Zero-Shot SOTA vs. our Fine-Tuned \datasetname Models.} The results demonstrate that fine-tuning significantly improves accuracy over zero-shot SOTA models.}
    \label{tab:model_comparison}
    \vspace{-3mm}
\end{table} \\
Our contribution aims to facilitate the development of  AI models that can handle complex cross-disciplinary questions in a truthful and comprehensive manner, enabling museums to serve as dynamic educational platforms that enrich visitor experience and deepen understanding across diverse cultural, historical, and scientific domains, as we show by fine-tuning BLIP \cite{li2022blip} and LLaVA.\cite{liu2023visualinstructiontuning} BLIP aligns images with descriptive text effectively, generating accurate captions that enhance its question-answering capabilities.
Still, BLIP’s smaller text encoder/decoder (\textit{BERT-base}, 110M params.) limits its ability to handle complex instructions. LLaVA, powered by the larger \textit{Llama-7B} LLM, excels in instruction comprehension and vision-language reasoning, making it capable of performing complex tasks. We provide insights into the nuanced and detailed understanding and real-world applications required for museum exhibits, presenting comparisons of the two models on multiple metrics.
We show both can handle questions well when answers can be directly derived from visual features. 
However, for questions requiring the integration of visual features with broader human knowledge, large VLMs attain higher accuracy, performing the reasoning needed for such inquiries. For instance, they can answer questions that link visual details to historical facts or explain connections to related events or figures not directly depicted.
The major contributions of the paper are:
\begin{itemize}[noitemsep,topsep=0pt]
\item \emph{Dataset and fine-tuned models:} We introduce a dataset of 65M images and 200M question-answer pairs for museum exhibits suitable to build new vision-language models and to fine-tune existing ones (e.g. BLIP, LLaVA) 
\item \emph{Benchmark:} We propose 5 tasks derived from our dataset, setting directions for research in real-world AI for cultural heritage, along with the metrics to evaluate them.
\item \emph{Results and insights:} We offer several insights about the collected dataset as well as the real-world tasks proposed. 
\end{itemize}
\vspace{-1 ex}

\begin{table}[]
\footnotesize
\vspace{0.5mm}
\resizebox{0.95\linewidth}{!}{
\begin{tabular}{lcrcc}
\toprule
Dataset                                                             & Domain                                                                            & \#images                & \#questions                                  & Public                                    \\
\midrule
Sheng et al. \cite{sheng2016dataset}    & Archaeology                                                                       & 160                     & 800                     & \xmark                     \\
AQUA \cite{garcia2020dataset}           & Art                                                                               & 21K                     & 80K                     & \cmark                     \\
iMet \cite{zhang2019imet}                & Art, History                                                                      & 155K                    & 155K                     & \cmark                     \\
VISCOUNTH \cite{becattini2023viscounth}  & Art                                                                               & 500K                    & 6.5M                     & \xmark                     \\
MUZE \cite{balauca2024taming}                 & Art, History                                                                     & 210K                    & 1.5M                     & \cmark                     \\
\midrule
\makecell{\datasetname\\(ours)}                                        & \makecell{Art, History,\\Nat. Sciences} & 65M                     & 200M                     & \cmark                     \\ 
\bottomrule
\end{tabular}}
\vspace{-2 ex}
\caption{\textbf{Literature comparison}. \datasetname v.s. related datasets from literature based on data domains, size and structure.}
\label{tab:rel_work}
\vspace{-6mm}
\end{table}

\section{Related Work}
\vspace{-1mm}
\label{sec:related}
\begin{figure*}
    \centering
    \includegraphics[clip, trim=0cm 0cm 0cm 0cm, width=0.97\textwidth]{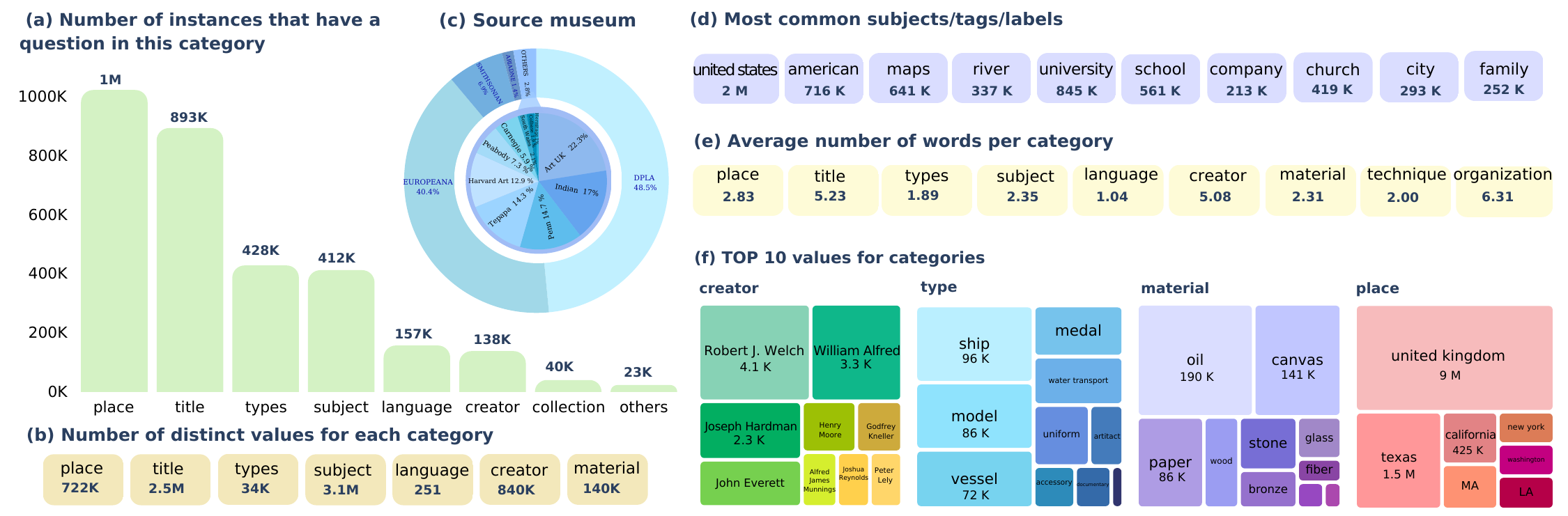}
    \vspace{-2.5mm}
    \caption{\textbf{Dataset statistics.} (a) distribution of questions, categorized by \textit{type}: the most common question is about the objects'\textit{ place of origin}, (b) number of distinct values of each category: the most varied category is \textit{subject} (c) data sources of each contributing museum, (d) the most common subjects/tags associated with the exhibits: objects coming from historical museums, like maps, items related to the United States, or personal themes, (e) average number of words (length) of each category value: \textit{organization} has the most words and (f) the most frequent values across different question categories: the objects' \textit{types} include ships, models, vessels, medals, and pieces of art.}
    \label{fig:donut}
    \vspace{-5.5mm}
\end{figure*}

\textbf{Vision language pre-training models and VQA.} 
Models like CLIP \cite{radford2021learning}, BLIP \cite{li2022blip} and LLaVA \cite{liu2023visualinstructiontuning}, pre-trained on large-scale datasets, have shown remarkable versatility in both unimodal and multimodal tasks~\cite{conde2021clip,chen2020uniter,kamath2021mdetr,li2019visualbert,li2020oscar,maaz2022class,lu2019vilbert,jia2021scaling,zhu2024llava}, incl. zero-shot recognition \cite{zhang2021tip,zhou2022learning,zhou2022conditional}, image segmentation \cite{ding2021decoupling,li2022languagedriven,zhou2022extract}, object detection \cite{gu2021open,bangalath2022bridging,zhou2022detecting}, etc. 
They offer a broad understanding of general concepts and can become valuable for specialized fields like cultural heritage and museums.
Previous studies on VQA have largely focused on images or videos, 
some works extending VQA by integrating external general knowledge~\cite{50wu2016ask,47wang2017fvqa,34marino2019ok} or knowledge tailored to 
specific datasets \cite{48wang2015explicit,12garcia2020knowit}. 
\\
\textbf{Digital humanities and cultural heritage.} In cultural heritage, achieving qualitative supremacy in visual understanding requires both informative images and reliable textual information. However, the required expertise is
a major challenge in data collection \cite{dataset2011novel,maji2013fine,wah2011caltech,sheng2016dataset,garcia2020dataset}. 
Multiple approaches for art understanding exist, including tasks such as cross-modal retrieval \cite{ananthram2023feelingblue}, image captioning \cite{bai2021explain,lu2022data,ruta2022stylebabel}, classifying \cite{ch16cetinic2018fine,ch39mensink2014rijksmuseum,ch40milani2021dataset,ch52tan2016ceci} or recognizing \cite{ch21del2019webly,ch30kahou2017figureqa} artworks. Previous attempts leverage existing cultural heritage data, approaching it from a multi-modal perspective~\cite{hannan2020manymodalqa,talmor2021multimodalqa,becattini2023viscounth,bai2021explain,lu2022data,gao2015you} but usually without using VLMs.
\\
MUZE \cite{balauca2024taming} achieves strong results on fill-in-the-gaps tasks by leveraging CLIP's multi-modal representations. However, its design relies on separate attention heads for individual attributes, making it both computationally expensive and challenging to scale for a large, diverse dataset like ours. Moreover, it does not align well with the direct Q\&A needs of our dataset, limiting its applicability to our tasks.
\\
\textbf{Domain-Specific datasets.}
General-purpose datasets \cite{34lin2014microsoft,22deng2009imagenet} are vast but lack domain-specific capabilities for cultural artifacts and scientific exhibits. For history and natural sciences~\cite{mensink2023encyclopedic, stevens2024bioclip}, datasets are scarce and often rely on external knowledge bases. In the Art domain, multiple datasets~\cite{ruta2022stylebabel, 52wilber2017bam, strezoski2018omniart} exist but mainly focus on artistic images with limited text and others~\cite{2achlioptas2021artemis, 20del2019noisyart, 5bianco2016predicting, 19ghosal2019aesthetic, 27garcia2020dataset, 16garcia2018read, 9bongini2020visual, 33malinowski2014multi, 23johnson2017clevr} combine visual and textual data but are either small, lack diversity, or rely on synthetic sources. VISCOUNTH~\cite{becattini2023viscounth} has 500K images and 6.5M questions only covering paintings and sculptures, while MUZE \cite{balauca2024taming} has 210K images and 1.5M texts in art and history (see \cref{tab:rel_work}). Our dataset of 65M images and 200M questions strikes a balance between scale and domain-specificity. It offers both the diversity and depth needed for a more comprehensive exploration of art, history and natural sciences VQA tasks, including data from museums used by previously mentioned works. We perform a benchmark comparison, evaluating the performance of our best BLIP model against BLIP trained on the MUZE dataset, showing that our dataset offers superior utility and effectiveness over existing alternatives with the experiment results being highlighted in \cref{app:benchmark_comparison}.

\vspace{-1.5 ex}

\section{Dataset}
\vspace{-1 ex}
\label{sec:dataset}
We built \datasetname, a multi-modal dataset containing 65M images and 200M question-answer pairs in multiple languages, ensuring cultural diversity, see \cref{fig:teaser}.

\subsection{Data Collection}
\vspace{-1 ex}
\datasetname covers 50M objects with questions in English and 15M with questions in \textbf{37 languages from Europe and Asia} (\textit{French, Spanish, German, etc}). See the list of all languages in \cref{app:multilingual_dataset}.
\datasetname is created by scraping museum websites of 3 prime international aggregators (DPLA, Europeana, Smithsonian), covering museums from Europe and North America and 12 other individual museums (see \cref{tab:raw_dataset} in \cref{app:dataset_details}) spread over the other continents.
Some museums consist multiple images of the same object  from different angles. We collected the web urls of all the images.
We show more details about the data origin in \cref{fig:donut}. We will make the dataset publicly available under the same license museums use, CCBY-NC-4.0.
\vspace{-0.3 ex}

\begin{figure*}
    \centering
    \includegraphics[clip, trim=0cm 0cm 0cm 0cm, width=1\textwidth]{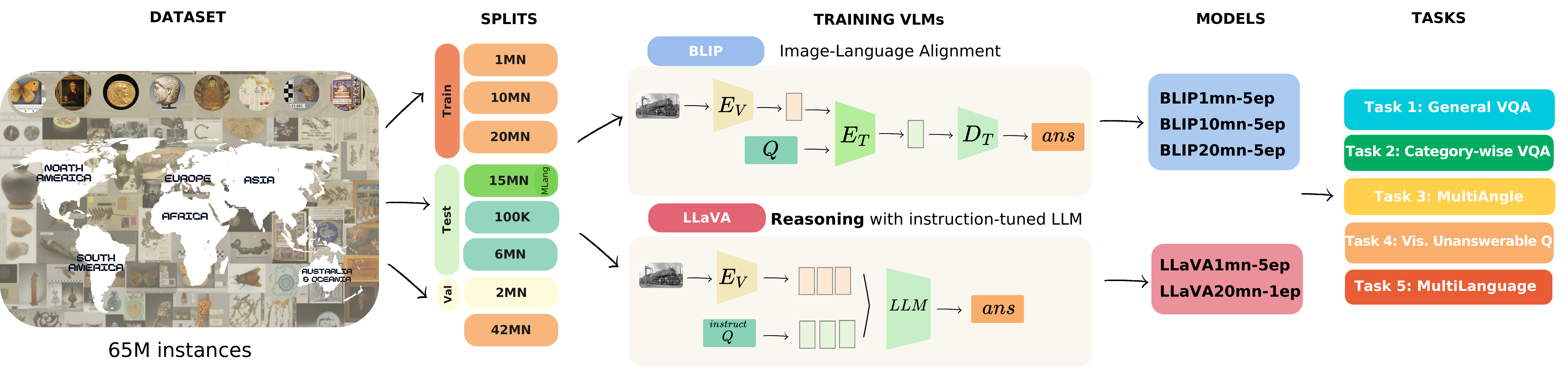}
    \vspace{-5mm}
    \caption{\textbf{Workflow}. Using smaller subsets of the dataset (1mn, 10mn and 20mn), we fine-tune BLIP and LLaVA models. \textbf{BLIP}, an encoder-decoder based model, \textbf{aligns language and image} in the same space while \textbf{LLaVA}, built on an instruction-tuned LLM is \textbf{directly reasons} based on the language.
}
    \label{fig:flow}
\vspace{-5mm}
\end{figure*}
\vspace{-0.3 ex}
\subsection{Data Curation}
\vspace{-1 ex}
A total of 10 experts worked over 3 months, 2 experts cross-checked for quality to clean and curate the entire data. 
\\
Tabular representation in the form of attribute-value pairs is the usual format for museum exhibit information. Each museum has a unique set of attributes. After extracting the attributes, we reformulate them as questions and associated values become answers. For detailed method, see Supp mat section 2.5. The overview of the method is as below. 
\\
{\bf Separating into attribute-value pairs.}
Information about exhibits takes 2 forms: (a) attribute-value pairs, scraped using museums APIs; (b) single strings, otherwise.
We determine separators to obtain the attribute-value pairs when object information is retrieved as a complete string.
\\
{\bf Filtering attributes.}
The object attributes also include \textit{display site in museum, catalog number, inventory date, dimensions}, and more. These are redundant for our goals and we excluded them from the main dataset. The remaining attributes were again divided into 2 types: (a) medium length attributes (with a length less than 100 words) (b) long length attributes, the rest. The reason is the restriction to 512 input tokens for BLIP. Despite LLaVA allowing for more input tokens, the final dataset on which our models have been trained was limited to the \textit{medium attributes}, thus ensuring a fair comparison of BLIP vs. LLaVA. 
When referring to our dataset in terms of training, validation or testing, we refer to the one with \textit{medium attributes} only. However, we will make the complete dataset along with the filtered and long-length attributes publicly available as the raw version.
\\
{\bf Creating questions from attributes.}
We structure attribute data for visual question answering separately for each museum, adapting to their format differences.
Questions are manually crafted (63 unique questions, listed in \cref{tab:category_questions} from \cref{app:question_list}) with attribute's value serving as answer. Humans formulated the questions to ensure diversity, having slightly varied questions for the same attributes across different museums, mimicking natural human curiosity to phrase questions in varied ways. For example, for the attribute \textit{material}, two varied questions were:
\textit{Which primary material is the object made of?} vs.
\textit{What is the material used in the object?}
\\
{\bf Creating the final dataset.}
We download all images from the collected image-urls. For each object, we now have a list of images and a set of question-answer pairs, omitting the answers for which the value is not known. Finally, for each museum we create 3 columns - image (having the list of images from different viewing angles), question (having the list of all questions), answer (having the list of respective answers).
Each question's answer is a list, since multiple answers may apply. See \cref{app:example_instance} for an example.
\vspace{-2 ex}
\subsection{Data Statistics and Bias Analysis}
\vspace{-1 ex}
We analyzed the dataset by examining question distribution, category diversity,  sources, common subjects, word counts per category, and frequent  question types (See \cref{fig:donut}). 
\\
While bias-free datasets are unattainable \cite{fabbrizzi2022survey}, we ensure \textbf{our dataset is bias-aware}. Our primary data sources, international aggregators, naturally emphasize European and American objects, leading to a \textbf{selection bias}, further amplified by the lack of digitization in smaller museums. Our dataset includes 5M+ objects from other continents. Nevertheless, results clearly show that finetuning on \datasetname causes benefits to distribute evenly despite regional biases (see \cref{tab:metrics-4}). Given the aggregators’ extensive curation, our collection spans a vast historical timeline, from ancient artifacts to modern art, covering statues, paintings, vessels, fossils, corals, war depictions, weapons, manuscripts, textiles, coins, and more. To mitigate \textbf{language bias}, we include 15M samples across 37 languages beyond English, with ongoing expansions. We also acknowledge \textbf{framing bias}, as models are trained on front-view images as per standard digitization practices, yet multi-angle experiments confirm model robustness to different image perspectives.
\\
To help researchers analyze and address biases, we will release MUSEUM-65 with tools for large-scale dataset exploration. These tools will enable image retrieval via text or image queries, aiding systematic bias detection and mitigation. By making the source code and essential routines publicly available, we aim to support customized dataset curation while fostering transparency and inclusivity. Additionally, we encourage users to explore the dataset and, in the future, report undetected biases and model behaviors through a planned public portal,  improving data curation and solidifying MUSEUM-65 as a real-world dataset.
\\
For applications requiring a minimally biased dataset, debiasing techniques such as model-agnostic training or specialized architectures will be commended \cite{yuan2021language, ouyang2021suppressing, gu2024beyond}.

\vspace{-0.5 ex}
\subsection{Societal Impact of Dataset}
\vspace{-1 ex}
Our dataset supports training multimodal models that enhance cultural accessibility, educational tools, and virtual heritage exploration, while promoting multilingual data and cross-cultural appreciation by enabling global artifact comparison.
Inspection of images and text reveals that museums, as reputable institutions, curate collections to address controversies—such as historical disputes, privacy, religious issues, and racial bias—and tag inappropriate content, ensuring dataset safety and quality. While origin bias remains a concern, we aim to mitigate it through collaborations and diversification, hoping broader museum digitization will further enhance diversity.
In the current form, we consider this dataset a research artifact and strongly advocate \textbf{academic use only} and advise careful investigation of downstream model biases (further analysis in \cref{app:dataset_bias}).
 
\begin{figure*}
    \centering
    \includegraphics[clip, trim=0cm 0cm 0cm 0cm, width=1\textwidth]{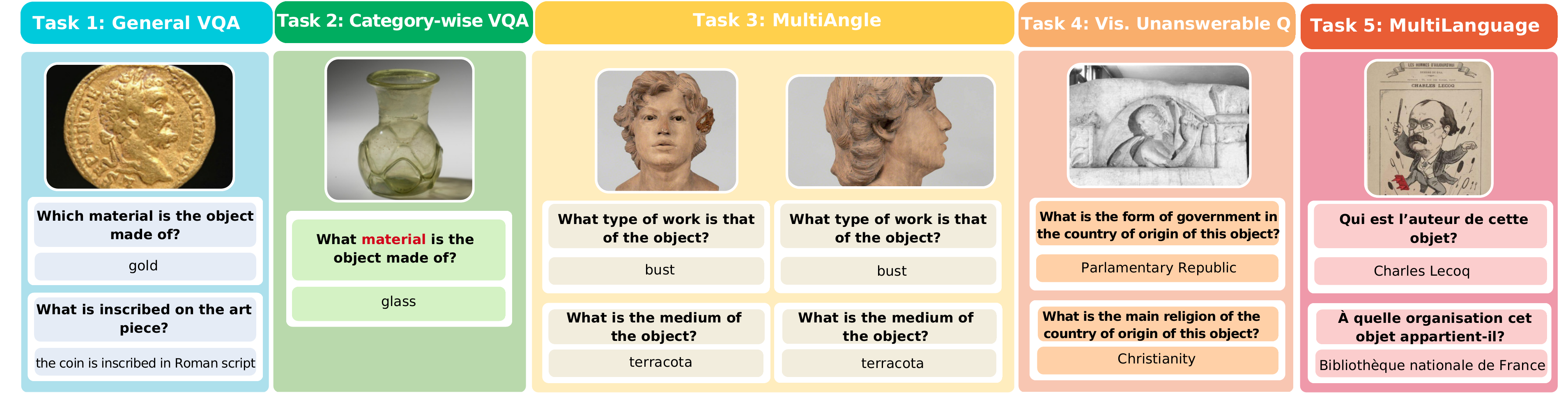}
    \caption{\textbf{Benchmarked tasks}. (1) \textbf{general VQA}, (2) \textbf{category-wise VQA}, (3) \textbf{MultiAngle} - measures the adaptability to different angle images of the same object, (4) \textbf{Visually Unanswerable Questions} - observes the response to new common knowledge questions derived from dataset's available information for an exhibit, (5) \textbf{MultiLanguage} - checks the ability to use languages like French and German} 
    \label{fig:tasks}
    \vspace{-5mm}
\end{figure*}


\vspace{-0.5 ex}
\subsection{Data Splits}
\vspace{-1 ex}
We split the data (English) in train, val and test, having 42M, 2M and 6M images, with an average of 3.5 questions per image (15M instances in other languages are in a separate test split). We create multiple smaller train subsets of 1M, 10M, 20M, and a smaller subset of the test dataset, with 10K instances, which we use during experiments and evaluation. The stratification is done to meet different computational needs. For more details about the splits, as well as the data format and examples, see \cref{app:data}.
\vspace{-1mm}
\section{Evaluation}
\vspace{-1 ex}
\label{evaluation}
We compute two types of metrics: (1) traditional uni-gram and n-gram-based numeric metrics that rely on lexical overlaps, and (2) deeper semantic-based metrics that leverage word embeddings for a more nuanced evaluation.

\noindent {\bf Setup.}
To ensure accurate and consistent metric calculations, we pre-process the answers by removing special characters, retaining only alphanumeric content before computing the metrics. The overall metric is an average of individual metric scores for each question.

\vspace{-0.5 ex}
\subsection{Numeric metrics}
\vspace{-1 ex}
We compute the commonly used \textbf{precision, recall, and BLEU scores}. To simplify evaluation, we introduce Complete Precision, which is the percentage of questions where the answer fully matches the ground truth (precision = 1.0). Similarly, Partial Precision is the percentage where there is at least some overlap (precision $>$ 0.0). Complete Recall and Partial Recall are defined analogously. The BLEU score \cite{papineni-etal-2002-bleu} measures the fraction of word n-grams in the model’s prediction that appear in at least one valid answer, with a brevity penalty to discourage short responses. We scale BLEU scores between 0 and 100, reporting results for BLEU1 (1-gram) and BLEU2 (2-gram). For detailed explanation of metrics, see \cref{evaluation_app}.

\vspace{-0.8 ex}
\subsection{Semantic metrics}
\vspace{-1 ex}
Some attributes like subject and short description, where textual variations in answers are equally valid, make numeric metrics insufficient. Semantic metrics offer a deeper evaluation of the model’s domain understanding by capturing contextual meaning rather than relying solely on exact word matching. Results (\cref{tab:semantic-metrics}) show that fine-tuning on \datasetname significantly improves these metrics.
\\
{\bf METEOR Score.}
METEOR aligns words using synonyms, stemming, and paraphrasing, making it more robust than pure n-gram overlap metrics. The final score considers precision, recall, and a fragmentation penalty to account for word order. We scale the score between 0 and 100 and average it across all instances.
\\
{\bf Word Mover's Distance Score.}
We also report Word Mover’s Distance (WMD) based top-1 accuracy \cite{kusner2015word}, which measures the minimum cumulative distance required to transform the predicted response into the ground truth in the Word2Vec embedding space. The most probable class is determined based on the smallest WMD score and accuracy of determining the ground truth class is calculated.
 \begin{table}[h!]
\centering
\vspace{-2mm}
\setlength{\tabcolsep}{3pt} 
\resizebox{0.75\linewidth}{!}{
\begin{tabular}{lcccc}
\toprule
Model              & \makecell{BLIP \\ nofinetune} & \makecell{BLIP \\ 20mn 5e} & \makecell{LLaVA \\ nofinetune} & \makecell{LLaVA \\ 20mn 1e} \\
\midrule
METEOR & 3.24 & \textbf{37.45} & 2.96 & \textbf{58.85} \\
WMD Acc. & 35.54 & \textbf{74.02} & 54.5 & \textbf{87.02}\\
\bottomrule
\end{tabular}%
}
\vspace{-3mm}
\caption{\textbf{Semantic Evaluation results.} Results demonstrate a substantial enhancement in domain understanding after fine-tuning}
\label{tab:semantic-metrics}
\vspace{-7mm}
\end{table}
\vspace{-0.3 ex}
\section{Experiments}
\vspace{-1 ex}
\label{sec:applications}
We introduce a comprehensive benchmark for \datasetname, evaluating general and specific tasks across different metrics by exploring multiple VQA-based tasks, including general question VQA, category-wise VQA, and three more challenging tasks designed to address real-world problems relevant to Museum LLMs. We also perform an ablation experiment on the place of origin to check if our models are biased to giving more accurate results to objects that belong from a specific region. This benchmark standardizes comparisons across methods, guiding future research toward effective models and identifying areas for improvement. See \cref{fig:flow} for an overview.
\vspace{-0.8 ex}
\subsection{Experimental Setup}
\vspace{-1 ex}
In our experiments we use two models known for VQA tasks, LLaVA \cite{liu2023visualinstructiontuning} and BLIP \cite{li2022blip} using our dataset. We train multiple model configurations with varying amounts of data and training epochs to analyze the impact of training time and data size on the results. We evaluate their performance using multiple scores (precision, recall, BLEU), and discuss their behavior. For further details and why we choose BLIP and LLaVA models see \cref{app:exp_details}.
\vspace{0mm} 
\begin{figure}[t]
    \centering
    \includegraphics[clip, trim=0.26cm 0.2cm 0.26cm 0cm, width=.5\textwidth]{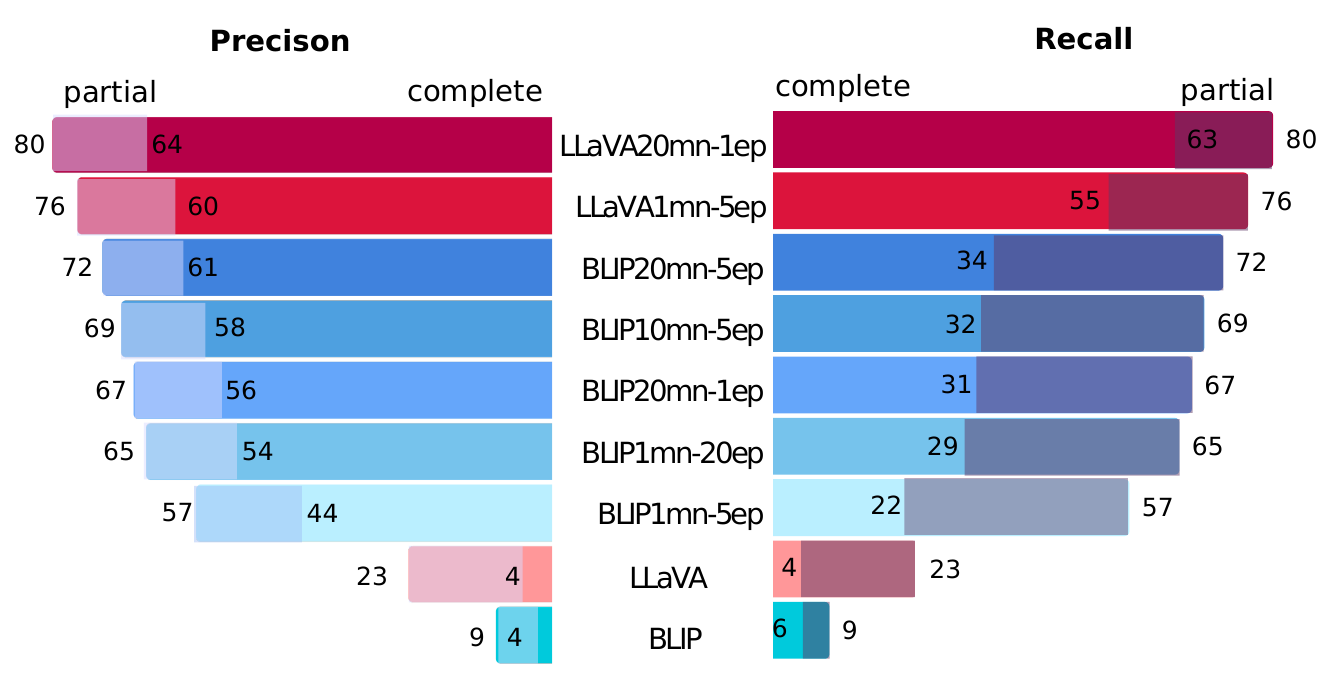}
    \vspace{-2mm}
    \caption{\textbf{General VQA results}. Comparison of fine-tuned and non-fine-tuned models on precision and recall. Models fine-tuned with the 20mn dataset perform best, with \textbf{LLaVA20mn-1ep} achieving 80\% partial precision and 64\% complete precision. \textbf{LLaVA models also outperform BLIP in recall}, indicating their predictions more often contain or are contained in the ground truth.}
    \label{fig:all_cmp}
    \vspace{-7mm}
\end{figure} 
\\
\vspace{0mm}\noindent{\bf Training on our dataset.}
We fine-tune LLaVA and BLIP using the same image-question pairs, choosing for every image one random question-answer pair every epoch. In each case, the front view image of an object is used.
\vspace{0.5mm}
\\
{\bf Finetuning BLIP.}
In our experiments we use BLIP, with the configuration available as \textit{blip-vqa}.
We fine-tune three main versions of BLIP, using: (a) 1mn train dataset for 5 epochs, extended up to 20 epochs (independently fine-tuned), (b) 10mn train dataset, 5 epochs, (c) 20mn train dataset, 5 epochs referring to them  as BLIP1mn-5ep, BLIP10mn-5ep, and BLIP20mn-5ep respectively. We also fine-tune a 20mn train dataset version for exactly 1 epoch to have a fairer comparison for LLaVA20mn-1ep. During fine-tuning we use a batch size of 512, mainly following the fine-tuning scheme of \cite{li2022blip}. More details in \cref{app:exp_details}.
\begin{figure}[t]
    \centering
    \includegraphics[clip, trim=0cm 0cm 0cm 0cm, width=0.499\textwidth]{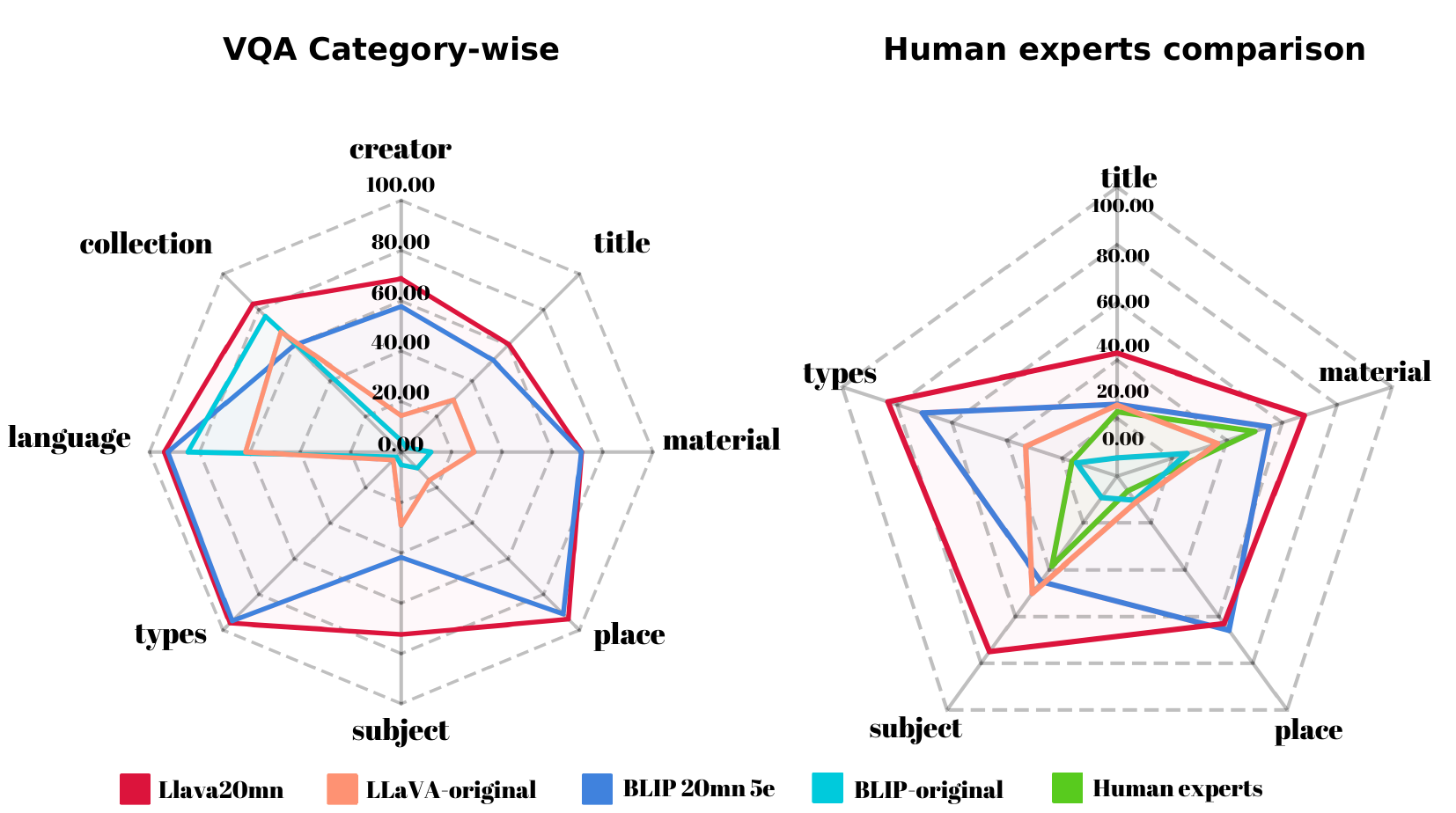}
    \caption{\textbf{VQA category-wise results.} Comparison of models with human experts (right) and fine-tuned vs. original models (left). \textbf{Fine-tuned models outperform in all categories}, \textbf{LLaVA20mn performing best}.  Fine-tuned models exceed human performance. The originals excel in \textit{language} and \textit{collection} due to common knowledge answers and fewer related instances.}
    \label{fig:category_wise}
    \vspace{-5mm}
\end{figure} \\
\noindent{\bf Finetuning LLaVA.}
For finetuning LLaVA, we assure the use of the same object-question pairs and the same order as for BLIP experiments.
We fine-tune two versions of LLaVA, (a) using 1mn train dataset for 5 epochs, and (b) using the 20mn dataset for 1 epoch. We will refer to them as LLaVA1mn-5ep and LLaVA20mn-1ep. We use a batch size of 512. We evaluate all models on the VQA tasks. \\
\noindent {\bf Hardware.} We train and evaluate our models using 64$\times$NVIDIA H100 GPUs.
\vspace{-1 ex}
\subsection{Task 1: VQA on general questions}

\vspace{-1 ex}
The task involves using all the questions associated with each image and producing the individual scores described in \cref{evaluation}. We compute the average score over all image-question pairs for each metric, to observe the model's general VQA capability and adaptability across a diverse range of visual and linguistic contexts, providing  the performance on any kind of question addressed by the user.
\\
While evaluating the fine-tuned LLaVA and BLIP on all the questions we observe that the LLaVA models are always receiving better results than their BLIP counterpart (See \cref{fig:all_cmp}). LLaVA20mn trained 1 epoch receives the best results having for 80\% of the predictions at least a part in common with the ground truth, and 63\% perfect match (prediction and ground truth are equal). We observe that the LLaVA models (fine-tuned 1mn or 20mn, and original LLaVA) have usually a close result between precision and recall, while the BLIP models (fine-tuned and original) have a big decrease in complete recall (the ground truth is completely present in the prediction). 
\begin{figure*}
    \centering
    \includegraphics[clip, trim=0cm 0cm 0.1cm 0cm, width=1\textwidth]{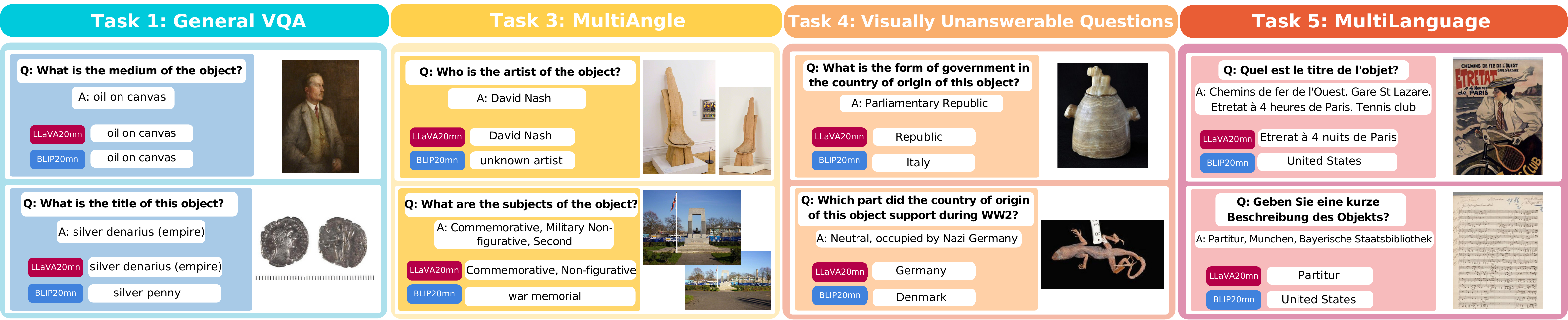}
    \vspace{-6mm}
    \caption{\textbf{Examples}. LLaVA20mn-1ep and BLIP20mn-5ep behaviour on different tasks, General VQA (1st column), MultiAngle (2nd column), Visually Unanswerable Questions (3rd column) and MultiLanguage (4th column). We observe \textbf{more precise answers for LLaVA20mn} than for BLIP20mn on all the tasks. Also the \textbf{last two tasks seem to be impossible for BLIP20mn}.}
    \label{fig:examples}
    \vspace{-3mm}
\end{figure*}

\vspace{-0.5 ex}
\subsection{Task 2: VQA category-wise}
\vspace{-1 ex}
Questions are grouped by attributes (eg: \textit{title, creator, technique, subjects/labels}). For each category, relevant questions are compiled (e.g asking about the title, denomination, or object name collected under \textit{title}). The model then answers each question, generating individual scores, which are then aggregated to compute an average score for each category, allowing for a detailed analysis of the model’s strengths and weaknesses across different categories, revealing areas where it may excel or struggle. We list all questions attributed to each category, in \cref{tab:category_questions}, \cref{app:question_list}.
\\
For this experiment, we compare the partial precision. We see in \cref{fig:category_wise} that LLaVA and BLIP original have very low results for most of the categories. We notice LLaVA fine-tuned having significantly better results than BLIP fine-tuned on \textit{subject} and \textit{collection}. The lowest result for all models are for title, which is also very difficult for humans.
\\
{\bf Human experts evaluation on VQA category-wise task. }
We randomly selected 850 question-answer pairs covering different attributes and conducted an experiment with 10 museum experts, who answered the same questions as our models. Their responses were evaluated across the categories \textit{types, title, place, material, and subject} using the same methodology as for the models. The results (see \cref{fig:category_wise}) reveal that certain categories, such as \textit{place} and \textit{types}, are particularly challenging for humans. Notably, fine-tuned models outperform human experts across all categories, especially in \textit{subjects, place, and types}, highlighting the need for specialized models with domain-specific knowledge. For \textit{materials}, performance is comparable, as these can be determined by simply observing the object.
\\
{\bf IAA metrics: }
We approximate Fleiss’s Kappa by simulating categorical behavior for free-form answers. Five experts independently answered a set of 63 unique questions (from test set). The “best” response was chosen via majority voting, and agreement was measured as proportion of the remaining four matching it, yielding \textbf{52.7\%} agreement—well above the \textbf{6.25\%$=0.5^4$} expected by chance.

\begin{table}[]
\footnotesize
\centering
\resizebox{.97\linewidth}{!}{
\begin{tabular}{l l ccccc}
\toprule
Model                                  & Angle       & \makecell{Partial \\ Prec.} & \makecell{Complete \\ Prec.} & \makecell{Partial \\ Recall} & \makecell{Complete \\ Recall} & {BLEU1} \\
\midrule
LLaVA20mn-1ep                          & Original    & 58.09                        & 46.09                        & 58.12                         & 41.04                         & 42.14   \\
                                        & Alternative & 56.14                        & 44.89                        & 56.15                         & 40.01                         & 41.02   \\
\midrule
LLaVA no finetune                      & Original    & 24.35                        & 0.09                         & 24.35                         & 11.25                         & 1.61    \\
                                        & Alternative & 23.56                        & 0.02                         & 23.56                         & 10.85                         & 1.54    \\
\midrule
BLIP-                           & Original    & 52.78                        & 42.51                        & 52.78                         & 35.29                         & 38.31   \\
                                        & Alternative & 51.75                        & 41.87                        & 51.75                         & 34.59                         & 37.62   \\
\midrule
BLIP nofinetune                       & Original    & 13.82                        & 9.70                         & 13.82                         & 5.22                          & 6.52    \\
                                        & Alternative & 12.86                        & 8.71                         & 12.86                         & 4.72                          & 5.92    \\
\bottomrule
\end{tabular}
}
\vspace{-2 ex}
\caption{\textbf{MultiAngle results}. Fine-tuned LLaVA20mn-1ep and BLIP20mn-5ep v.s. no fine-tune models.  \textbf{The models are stable w.r.t. the viewpoint changes}. (Please refer to \cref{fig:examples}-3rd col.).}
\vspace{-5mm}
\label{tab:multiangle}
\end{table}

\vspace{-0.5 ex}
\subsection{Task 3: Multi Angles}
\vspace{-1 ex}
To assess the model’s resilience to viewpoint changees, we evaluate it using images captured from different angles or perspectives, available in our dataset. By substituting these viewpoint-varied images for the originals, we can directly compare these scores with those from the initial baseline images to observe any shifts in accuracy or relevance. 
\\
For this task we select a subset of $\approx$ 5K exhibits from the test dataset with multiple images taken from different angles (e.g. 2nd column of \cref{fig:examples}). In total we evaluate on $\approx$ 22K questions. All our models (\cref{tab:multiangle}) show consistent scores when presented with images from different angles, suggesting a strong capacity for generalization and an ability to recognize objects despite variations in angle or orientation, providing insights into the model’s ability to maintain performance stability when faced with real-world variability in image capture. The slight performance drop can be attributed to a decrease in image information (e.g. pictures of statues from the side are generally harder to recognize).

\vspace{-0.5 ex}
\subsection{Task 4: Visually Unanswerable Questions}
\vspace{-1 ex}
We introduce a set of specialized questions to assess the model’s contextual understanding, focusing on an object's country of origin or creator. These carefully designed questions require a deeper level of contextual or associative reasoning. For example, instead of simply asking about characteristics that may be linked with a visual pattern (assuming that the painters' style can be visually recognized - \textit{``Who is the painter of this painting?''}), these questions may ask, \textit{``Who was the mentor of the painter of this painting?''} or \textit{``What is the nationality of the painter of this painting?''}.  
\\
We manually generate 5-6 questions for exhibits, related either to the creator or country and search for answers online (e.g. 3rd column of \cref{fig:examples}). We obtain 510 and 515 image-question-answer pairs from the train and test dataset respectively. This approach evaluates not only whether the model can correctly identify or infer the country of origin or creator based on visual cues but also tests its ability to correlate these features with general knowledge or cultural information, addressing beyond surface-level visual details. The full list of questions is available in \cref{tab:task4Q} from \cref{app:vuq_questions}. According to results in \cref{tab:nonvisq_train}, both original and fine-tuned LLaVA have much higher reasoning capabilities than BLIP, due to LLaVA's higher model size and larger pre-training dataset. Moreover, fine-tuning LLaVA enhances its ability to reason about museum exhibits, esp. when considering the precision of its answers. On the other hand, BLIP's performance on this complex task drops after fine-tuning, hinting at BLIP's limited model capacity causing forgetting of prior knowledge in order to accommodate the new training data. The consistency of results across the test dataset further supports LLaVA20mn-1ep’s ability to reason beyond visual features even on unseen images (see \cref{tab:nonvisq_test}).
\begin{table}[]
\vspace{-0.5mm}
\centering
\resizebox{0.92\linewidth}{!}{
\begin{tabular}{lccccc}
\toprule
Model              & \makecell{partial \\ prec.} & \makecell{complete \\ prec.} & \makecell{partial \\ recall} & \makecell{complete \\ recall} & {BLEU1} \\
\midrule
LLaVA20mn-1ep      & \textbf{31.37}                          & \textbf{25.1}                           & \textbf{31.37}                       & \textbf{12.94}                       & \textbf{15.16}                                \\
LLaVA no finetune              & 24.27                         & 0.58                           & 24.27                       & 6.21                        & 1.74                             \\
BLIP 20mn-5ep & 2.35                          & 0.2                            & 2.35                        & 0.2                         & 0.63                          \\
BLIP no finetune    & 6.08                          & 5.69                           & 6.08                       & 2.75                        & 2.95                  \\   
\bottomrule
\end{tabular}
} 

\vspace{-1.5 ex}
\caption{\textbf{Visually Unanswerable Questions results on train images}. Clearly, \textbf{LLaVA20mn-1ep}, performs best, especially for complete precision and complete recall, showing the \textbf{ability} to visually link the objects with the corresponding dataset information and \textbf{to respond to visually unanswerable} \textbf{questions}.}
\vspace{-0mm}
\label{tab:nonvisq_train}
\end{table}

\begin{table}[t]
\centering
\vspace{-3.5mm}
\resizebox{0.92\linewidth}{!}{
\begin{tabular}{lccccc}
\toprule
Model              & \makecell{partial \\ prec.} & \makecell{complete \\ prec.} & \makecell{partial \\ recall} & \makecell{complete \\ recall} & {BLEU1} \\
\midrule
LLaVA 20mm-1ep & \textbf{29.7} & \textbf{25.83} & \textbf{29.7} & \textbf{10.29} & \textbf{12.67} \\
LLaVA nofinetune & 27.18 & 1.55 & 27.18 & 6.41 & 3.08 \\
BLIP 20mm-5ep & 3.3 & 0.78 & 3.3 & 0.19 & 0.73 \\
BLIP nofinetune & 5.44 & 5.24 & 5.44 & 2.33 & 2.58 \\
\bottomrule
\end{tabular}%
}
\vspace{-3mm}
\caption{\textbf{Visually Unanswerable Questions results on test images.} Results are consistent with those of the train split indicating the capability of the model to generalise well on unseen images.}
\vspace{-6mm}
\label{tab:nonvisq_test}
\end{table}

\vspace{-0.8 ex}
\subsection{Task 5: Multiple Languages}
\vspace{-1 ex}
We evaluate the model’s zero-shot performance on non-English questions, including \textit{French, German, Spanish}, and others available in the \textit{multilanguage} dataset split. Questions are formulated in the respective languages using collected attributes. This assesses the model’s ability to link visual content with multilingual queries, recognizing objects, actions, or scenes without relying on English training biases, which is crucial for real-world multilingual use.
\\
Lastly, we evaluate our models on 500 images with textual data in French and German, for a total of 2864 question-answer pairs (e.g. in \cref{fig:examples} - 4th column). In \cref{tab:multilang} we can observe that both variants of LLaVA achieve better results than BLIP. However, our fine-tuned LLaVA seems to have partially forgot its abilities to answer in foreign languages due to it being only fine-tuned with english data. Although the original LLaVA easily answers questions in different languages (it has high partial precision and recall), it mostly fails to give perfect answers. Further fine-tuning the models using multilingual data from \datasetname{} should improve their performance. A small-scale dataset was curated for Tasks 4 \& 5 to ensure quality, given the significant manual effort. We plan to continue scaling this curation.

\begin{table}[]
\vspace{-0.5mm}
\resizebox{1\linewidth}{!}{
\footnotesize
\centering
\begin{tabular}{lcccccc}
\toprule
                   \multirow{3.4}{*}{Model}
                   & \multicolumn{2}{c}{French}                                                                                          & \multicolumn{2}{c}{German}                                                                                          & \multicolumn{2}{c}{Average}                                                                                                                                                                                           \\
\cmidrule(rl){2-3} \cmidrule(rl){4-5} \cmidrule(rl){6-7}              & \makecell{partial \\ prec.} & \makecell{complete\\ prec.} & \makecell{partial \\ prec.} & \makecell{complete\\ prec.} & {BLEU1} & {BLEU2} \\
\midrule
LLaVA20mn-1ep      & {10.37}                                & {0.54}                                 & {9.72}                                 & {\textbf{1.17}}                                                                                                           & 1.36                      & \textbf{0.27 }                     \\
LLaVA nofinetune   & {\textbf{41.81}}                                & {0.4}                                  & {\textbf{18.41}}                                & {0.15}                                                                                                          & \textbf{1.46 }                     & 0.13                      \\
BLIP20mn-5ep & 4.02                                                       & 0.4                                                       & 0.73                                                        & 0.15                                                                                                                                  & 0.21                      & 0.01                      \\
BLIP nofinetune    & 2.01                                                        & \textbf{0.6}                                                        & 0.8                                                        & 0.29                                                                                                                                    & 0.13                      & 0  \\ \bottomrule                      
\end{tabular}
}
\vspace{-3mm}
\caption{\textbf{Multi-Language results}. LLaVA models perform better than BLIP ones. LLaVA20mn-1ep \textbf{slightly forgets the ability to answer in other languages}, due to its fine-tuning in English. However, on complete precision and BLEU2 the results of LLaVA20mn-1ep are sligthly better than the no fine-tune versions. }
\vspace{-6mm}
\label{tab:multilang}
\end{table}

\vspace{-0.5 ex}
\subsection{Place of Origin Ablation}
\vspace{-1 ex}
We curated 1K images per continent and evaluated our best models on it in Tab.~\ref{tab:metrics-4}. Despite the bias in place of origin, the benefits distribute evenly.

 \begin{table}[h!]
\centering
\vspace{-3mm}
\setlength{\tabcolsep}{3pt} 
\resizebox{0.9\columnwidth}{!}{
\begin{tabular}{lcccccc}
\toprule
Model & Europe & N. America & S. America & Asia & Africa & Oceania \\
\midrule
LLaVA 20mn-1ep& \textbf{85.2} & \textbf{79.6} & \textbf{86.6}  & \textbf{67.4} & \textbf{86.7} & \textbf{99.2} \\
LLaVA & 8.6 & 43.57 & 20.3 & 23.4 & 20.79 & 52.4\\
BLIP-20mn-5ep & 79.1 & 73.1 & 76.4 & 65.5 & 76.4 & 49.7\\
BLIP & 4.3 & 15.2 & 19.7 & 9.3 & 19.7 & 6.6\\

\bottomrule
\end{tabular}%
}
\vspace{-3mm}
\caption{\textbf{Continent-wise Partial Precision.} Despite of training data imbalance, the training on our dataset benefits all continents. }
\vspace{-6mm}
\label{tab:metrics-4}
\end{table}

\vspace{-1 ex}
\section{Conclusion}
\label{sec:conclusion}
\vspace{-1 ex}
We present a large, specialized dataset for VQA on museum exhibits, designed to bridge visual content and text-based queries. This dataset encompasses millions of images paired with varied questions, enabling models to deliver answers about a broad range of cultural artifacts. We fine-tune two VLMs, BLIP and LLaVA, to compare their performance on this museum VQA dataset. LLaVA, in particular, excels at answering visually unanswerable questions through reasoning and general knowledge. Additionally, cross-lingual tests confirm the adaptability of these models in multilingual contexts, highlighting their potential for use in diverse cultural and linguistic settings. This dataset and our experiments open doors for future applications in museum experiences. Models trained on \datasetname could support interactive virtual tours, where users ask detailed questions in their own languages. They could power digital curators, providing rich cultural insights, or integrate with AR to offer real-time, on-site interpretation, creating immersive learning experiences for museum visitors globally.



\section*{Acknowledgements}
We highly appreciate Pratyush Sinha, Krishnav Bajoria, Mohit Sharma, Anshuman Biswal, Rishabh Varshney, Anjali Roy, Raluca Mocanu, Reni Paskaleva and Nora Paskaleva for their help in gathering and curating the data, and for all the support, ideas and relevant discussions during the project. This research was partially funded by the Ministry of Education and Science of Bulgaria (support for INSAIT, part of the Bulgarian National Roadmap for Research Infrastructure). We thank the Bulgarian National Archaeological Institute with Museum for the support and guidance. We thank all institutions included in Europeana, Digital Public Library of America (DPLA), Smithsonian Institution, Ariadne Project and also to the aggregators themselves for providing open access to their data. We also thank to Carnegie Museums of Pittsburgh, Modern and Contemporary Art Museum Korea, Harvard Museums US, Peabody Museum US, ArtUK Project, Hermitage Museum Russia, South Wales Museum Australia, The Indian Museum Project, Colbase Project Japan, The Museum of New Zealand Te Papa Tongarewa and Penn Museum US for the access to their data that made this research possible. We thank Google DeepMind which provided vital support and resources for this research.

{
    \small
    \bibliographystyle{ieeenat_fullname}
    \bibliography{main}
}

\clearpage
\appendix
\section{Index}
\vspace{-3 ex}
\begin{table}[!h]
\centering
\resizebox{0.9\linewidth}{!}{
\begin{tabular}{cl} \toprule
\multicolumn{1}{l}{Section} & Section Name                                \\ \midrule
\textbf{1}                                  & \textbf{Index}                               \\
\textbf{2}                                  & \textbf{Data}                                \\
2.1                                & Data format                 \\
2.2                                & Example of instance                 \\
2.3                                & Data splits                     \\
2.4                                & Dataset details    \\
2.5 & Dataset Curation Process \\
2.6                                &     List of questions category-wise                      \\
2.7                                & Category analysis                   \\
2.8                                & Multi-lingual dataset
                           \\
2.9                            & Bias in dataset                         \\
2.10                            & Safety and Ethical considerations                         \\
\textbf{3}                                  & \textbf{Experimental details }             \\
3.1                                & Implementation details              \\
3.2                                & Evaluation metrics\\
3.3                                & Finetuning details                  \\
3.4                                & Training evolution                  \\
\textbf{4 }                                 & \textbf{Additional results  }                \\
4.1            & Benchmark Comparison         \\
4.2                                & General VQA                         \\
4.3                                & Category-wise VQA                   \\
4.4                                & MultiAngle VQA                      \\
4.5                                & Visually Unanswerable Questions VQA \\
4.6                                & MultiLanguage VQA                   \\
\textbf{5 }                                 &\textbf{Limitations and society impact}  \\ \bottomrule   
\end{tabular}}
\vspace{-2mm}
\caption{The index showing the additional information, technical details and results.}
\label{tab:index}
\end{table}
\vspace{-4 ex}
\section{Data}
\label{app:data}
\vspace{-1 ex}
This section provides comprehensive details about the dataset used in the task. It includes information on the raw dataset, an example of an instance, and the data format. Additionally, it outlines a category-wise list of questions, data splits, and a detailed category analysis, offering insights into the structure and distribution of the data.

\subsection{Data format}
\vspace{-1 ex}
All this curated information was stored in the form of json files in a dictionary format. With the object\_id being the key and the information in the respective value. 
\subsection{Example of instance}
\label{app:example_instance}
\vspace{-1 ex}
A detailed example from the dataset, showcasing the structure of an individual data point to clarify how the data is organized and used in the task is looking as follows:
\vspace{-2mm}
\begin{table}[!h]
\begin{tabular}{ll}
Question & \begin{tabular}[c]{@{}l@{}}{[}``Who is the artist of the object?'', \\ ``What materials is the object made of?''{]}\end{tabular} \\
Answer   & {[}{[}``Leonardo Da Vinci''{]}, {[}``wood'',``iron''{]}{]} \\
Image    & {[}``object1\_1'', ``object1\_2'', ``object1\_3''{]}       
\end{tabular}
\vspace{-2mm}
\end{table}
\subsection{Data splits}
\vspace{-1 ex}
This section details the dataset splits, including multiple training datasets designed to analyze the impact of varying data sizes. It also covers the validation split and multiple testing splits, enabling more efficient evaluation and comparison by reducing time requirements. The 42M train set is the original training set that we were able to collect, still due to time and other resources constraints we choose to fine-tune up to the 20M instances dataset.
\vspace{-1 ex}
\begin{table}[!h]
\small
\centering
\begin{tabular}{lcc}
\hline
Dataset                & Objects     & Q-A pairs \\ \hline
1mn\_train & 1M           & 3M  
\\
10mn\_train                       & 10M & 31M\\
20mn\_train                       & 20M &        61M                        \\
42mn\_train                       & 42M          & 123M                               \\
val        & 2M          & 4M                               \\

test                              & 6M           & 18M   \\
tiny\_test                        & 10K              & 30K\\
small\_test                       & 100K              & 3M\\    
multilingual & 15M& 45M
\\ \bottomrule     
\end{tabular}
\vspace{-2mm}
\caption{Description the dataset splits, including multiple training sets, a validation set, and several test sets. The splits are designed to facilitate analysis of performance under different training scenarios and streamline evaluation across various testing conditions.}
\label{tab:splits}
\end{table}

\begin{table*}[!h]
\footnotesize
\resizebox{\linewidth}{!}{
\begin{tabular}{crrrll}
\toprule
Museum Name  & \multicolumn{1}{l}{\#attributes} & \multicolumn{1}{l}{\#objects} & \multicolumn{1}{l}{\#images} & Trainable attributes                                                        & Non-trainable attributes                                                           \\ \midrule
Europeana    & 7                                        & 19163199                              & 23395805                             & \makecell[l]{organization, subject, type, country, \\title, creator}                        & description                                                                                                            \\ \midrule
Carnegie     & 6                                        & 76655                                 & 76655                                & creator, classification, credit, medium                                     & nationality, date                                                                                                      \\ \midrule
Contemporary & 3                                        & 9582                                  & 9582                                 & artist, title                                                               & date                                                                                           \\ \midrule
Harvard      & 9                                        & 579148                                & 265555                               & \makecell[l]{technique, classification, worktypes, \\century, medium}                       & division, creditline, department, period                                                                                           \\ \midrule
Peabody      & 9                    & 77379                                 & 77379                                &     \makecell[l]{title, material, place of origin, artist,\\ category, department, subjects,\\keywords associated,  short description}                                                                        & NA\\ \midrule
ArtUK        & 22                                       & 292358                                & 579148                               & tags, artist, title, medium, worktypes                                      & \makecell[l]{Acquisition method, Work status, Access note,\\Date Listing date, Installation end date,\\ Signature/marks description, Venue, Access, \\ Listing, Measurements, status, Unveiling date, \\Accession number, Installation start date, \\ Custodian, Inscription description,  Owner}                                                                                                                                     \\ \midrule
Hermitage    & 22                                       & 12572                                 & 14135                                & \makecell[l]{technique, school, place, title, author, \\material, epoch, category}          & \makecell[l]{Place of creation, Date, Inventory Number, \\Subcollection, Acquisition date, Dimension, \\ Place of finding, Collection, Complex., firm, \\ Manufacture, workshop, "Book, album, seria",\\Information about the original, \\Archaeological site, Comment}                                                                                                                                                              \\ \midrule
SouthWales   & 6                                        & 27433                                 & 46380                                & title                                                                       & \makecell[l]{exhibition history, audio, provenance,\\ video, places}                                                                                                                 \\ \midrule
Indian       & 34                                       & 189838                                & 313962                               & \makecell[l]{language, coin description observe, \\main material, main artist, inscription} & \makecell[l]{Accession Number, Artist Nationality, Mint \\ Title, Weight, Manufacturing Technique, Script,\\Historical Note, Detailed Description, Medium, \\Provenance, Museum Name, Patron Dynasty, \\Coin Description Reverse, Dimensions,\\ Find Place, Origin Place, Tribe, School,\\ Gallery Name, Title2, Number of Illustrations,\\ Brief Description, Subject, Scribe, Culture, \\Artist Life Date, Number of folios, Country} \\
\midrule
DPLA         & 6                                        & 22984790                              & 22984790                             & \makecell[l]{language, publisher, collection title, title,\\ place of origin, subject}      & NA                                                                                                                                                                                                                                                                                                                                                                                                          \\ \midrule
Colbase      & 14                                       & 22196                                 & 22196                                & category, genre, material, artist, holder                                   & \makecell[l]{Period/Century, Country/Origin, Donor, \\ Quantity,  Inscriptions, Excavation site, \\ Cultural property designation, Size,\\ Collection reference no.,}                                                                                                                             \\ \midrule
Tepapa       & 6                                        & 187595                                & 251361                               & collection, title, type, additionalType                                     & Caption, CreditLine                                                               \\ \midrule
Penn         & 12                                       & 191831                                & 556092                               & \makecell[l]{culture, culture area, continent, materials, \\technique, credit line, place}   & Description, length, width, height, depth                                                                          \\ \midrule
Smithsonian  & 4                    & 3277593                               & 3277593                              &         name, sex, place of origin, taxonomy                                                                    &    NA                                                                                                                \\
\midrule
Ariadne      & 4                                        & 665289                                & 665289                               & title, nativesubject, place                                                 & description \\ \bottomrule
\end{tabular}}
\vspace{-2mm}
\caption{The list of museums and aggregators. We display the number of attributes each museum have, the number of objects that they provided and the number of images available for them. We also present the attributes that helped the creation of the questions used during training and testing (Trainable attributes) as well as the attributes not used for questions but that we make available in the raw dataset (Non-trainable attributes).}
\label{tab:raw_dataset}
\vspace{-2mm}
\end{table*}
\vspace{-2mm}

\begin{table}[!h]
\footnotesize
\centering
\resizebox{0.87\linewidth}{!}{
\begin{tabular}{ll}
\toprule
Category        & Question                            \\ \midrule                         
Subject         & what are the subjects that the object depicts?                 \\
                & what are the subjects that are depicted by the object?         \\
                & which category does this object belong to?                     \\
                & what is the subject of this image?                             \\
                & what tags can the object be associated with?                   \\
                & under what category does this object fall?                     \\
                & what are the keywords associated with objects?                 \\
                & what is the category of the object?                            \\
                & what category does this object fall into?                      \\
                & what are the subjects of object ?                              \\ \midrule
Creator         & who is the publisher of this object?                           \\
                & who is the holder of the object?                               \\
                & who is the creator of the object?                              \\
                & who has created this object?                                   \\
                & who is the author of the text?                                 \\
                & who is author of the object?                                   \\
                & to whom is this object credited to?                            \\
                & who is the artist of the object?                               \\
                & who created this art?                                          \\ \midrule
Title           & what is the title of the object?                               \\
                & what is the name of the object?                                \\
                & what is the title of this object?                              \\
                & what is the name of the costume?                               \\
                & what is a suitable title for the object?                       \\
                & what is the denomination of the coin?                          \\
                & what can be the title of the object?                           \\
                & what is the title of the object                                \\ \midrule
Material        & which primary material is the object made of?                  \\
                & what material is the object made of?                           \\
                & what materials is the object made of?                          \\
                & which secondary material is the object made of?                \\
                & what is the medium used to create this object?                 \\
                & which tertiary material is the object made of?                 \\
                & what are the materials that this object is made up of?         \\
                & what is the medium of the object?                              \\ \midrule
Type            & which type of object is this?                                  \\
                & which type of object is it?                                    \\
                & what is the genre of this object?                              \\
                & what type of work is that of the object?                       \\
                & what is the additionaltype of the object?                      \\
                & what is the type of the object?                                \\ \midrule
Place & what is the place of origin of the object?                     \\
of Origin       & what is the place of origin of this object?                    \\
                & which country does this object belong to?                      \\
                & which continent does this object belong to?                    \\
                & what place could this object be from?                          \\ \midrule
Collection      & from which collection has this object been taken?              \\
                & what is the collection of the object?                          \\
                & what department does this object fall into?                    \\
                & what school does object belongs to?                            \\ \midrule
Technique       & what technique is used to make the object?                     \\
                & \makecell[l]{what is the technique that \\has been used to make this object?}  \\
                \midrule
Culture         & \makecell[l]{which area does the culture depicted\\ by this object belong to?} \\
                & which culture does this object belong to?                     \\
                \midrule
Language        & which language is the text in the object?                      \\
                & what is the language of the text?                              \\ \midrule
Others          & what is the object about?                                      \\
                & which period does this object belong to?                       \\
                & which style do the costumes belong to?                         \\
                & what is inscribed on the art piece?                            \\
                & what is the obverse of the coin?                               \\
                & which organization does this object belong to?           \\ \bottomrule     
\end{tabular}}
\caption{The questions generated from the attributes available for the exhibits grouped by categories.}
\label{tab:category_questions}
\vspace{-10mm}
\end{table}
\subsection{Dataset details}
\label{app:dataset_details}
We provide an overview of the dataset origin \cref{tab:raw_dataset}, including its composition, sources, and initial structure before processing. It highlights the foundational data used to create the final dataset for the task. We also show the amount of objects, images and attributes available from each museum, highlighting the attributes used for fine-tuning (Trainable attributes). The raw dataset will also be made publicly available along the curated dataset and it will also include the attributes not used for fine-tuning (Non-trainable attributes). The links to the curated and raw datasets can be found here: \href{https://github.com/insait-institute/Museum-65}{MUSEUM-65}
.

\subsection{Datasest Curation Process}
The dataset curation was done in 5 major steps:

\noindent {\it 1. Museum selection:}
\begin {itemize}
    \item The dataset comprises 65 million data points, with 95\% sourced from three major cultural aggregators: Digital Public Library of America - DPLA (24M), Europeana (20M), and the Smithsonian Institution (3.5M).
    \item These aggregators provide access to extensive digitized collections from major museums across Europe and America and offer structured data through platform-specific APIs.
    \item DPLA and Smithsonian provide metadata in English, whereas Europeana includes metadata in English as well as several European languages such as French, Spanish, and German.
    \item To ensure broader diversity in geography, culture, variety, and language, we curated the remaining 5\% of the dataset from 12 additional major museums spanning multiple continents. These additional museums were selected based on their global prominence and the richness of their collections, with data acquired through a custom scraping pipeline.
    \item Depending on the museum’s web infrastructure, data was collected either using official APIs or through HTML parsing tools such as BeautifulSoup.
    \item In cases where museums provided multiple images per object, we collected the URLs of all available views to preserve multi-angle visual representations.
    \item All this information was stored in the form of json files in a dictionary format. With the object id being the key and the information in the respective value.

\end {itemize}
\vspace{-0.5mm}

\noindent{\it 2. Data Cleaning:} Each museum's data was curated by a single domain expert to ensure consistency. Curation involved minimal edits: removing redundant attributes (inventory numbers, bibliographic info); extraneous symbols and numbers. Given high quality of museum data, focus was on consistent formatting rather than content rewriting.
\vspace{-0.5mm}
\noindent {\it 3. Attribute-Value Structuring:} While some museums provided such structured data, others required parsing complete strings, with experts identifying logical separators and attribute boundaries through example-driven consensus.
\vspace{-0.5mm}
\newline\noindent {\it 4. Question Crafting:} 
\begin{itemize}
    \item To structure the attribute–value data for the visual question answering (VQA) task, we aligned our approach with natural human curiosity—formulating questions and expecting concise answers—toward our goal of real-time deployment in interactive museum environments.
    \item Each museum’s data was processed independently due to differing metadata formats and schema structures.
    \item Experts manually authored natural language questions for a total of 63 unique attributes, with the corresponding attribute
    value serving as the ground-truth answer.
    \item Human synthesis ensured that even when the same attribute appeared across different museums, the phrasing of the questions varied to maintain linguistic diversity. Example: Both the questions {\it “Which primary material is the object made of?”} and {\it “What is the material used in the object?”} are related to the attribute "material"
    \item After synthesis, a centralized review process was conducted to remove redundancy, normalize structure where needed, and ensure phrasing diversity.
    \item We computed an average intra-category question similarity of 75\%, indicating a desirable balance between consistency and variation across museums and attributes.
\end{itemize}
\vspace{-0.5mm}
\noindent\textbf{\it 5. Final Assembly:} We download all the images from the collected image-urls.
For each object, we now have a list of images and a set of question-answer pairs, omitting the answers for which the value is not known. Finally, for each museum we create 3 columns - image (having the list of images from different viewing angles), question (having the list of all questions), answer (having the list of respective answers).
The answer to every question is in the form of a list as sometimes there may be multiple answers.
\vspace{-0.5mm}
\newline \noindent \textbf{Quality Control}: A data processing protocol covering data cleaning, consistency norms, and question design was shared with experts. Edge cases were discussed collaboratively, and the final dataset was schema-validated programmatically. 

\subsection{List of questions category-wise}
\label{app:question_list}
\vspace{-2mm}
We provide the categorization of the questions in the dataset. The questions are grouped based on their type or theme for an easier analysis during the evaluation. The \cref{tab:category_questions} is showing all these questions and their categories for a better understanding of the diversity of information and the variety of asking a question included in our dataset.

\subsection{Category analysis}
\vspace{-1 ex}
We present in \cref{tab:analysis} the top values and their frequencies across various categories, providing insights into the most prominent features and trends within the dataset.

\subsection{Multilingual Dataset}
\label{app:multilingual_dataset}
\vspace{-1 ex}
Our multilingual dataset comprises 15 million datapoints, featuring objects described in 37 European and Asian languages. \Cref{multi_ling_disribution} shows the distribution of exhibits across these languages:
\vspace{-1 ex}
\begin{table}[h]
\centering
\resizebox{0.87\linewidth}{!}{
\begin{tabular}{lclcll}
\hline
Language  & \%    & Language  & \%    & \multicolumn{2}{c}{Other Languages (\%)} \\ \midrule
German    & 16.62 & Italian   & 2.57  & Lithuanian  & \multirow{7}{*}{21.99} \\ 
Norwegian & 11.89 & Polish    & 2.56  & Romanian    \\ 
Dutch     & 11.79 & Estonian  & 2.41  & Croatian    \\ 
Spanish   & 8.80  & Czech     & 1.54  & Portuguese  \\ 
French    & 7.96  & Finnish   & 1.16  & Bulgarian   \\ 
Swedish   & 5.65  & Catalan   & 1.12  & Greek       \\ 
Danish    & 3.10  & Hungarian & 0.84  & and more    \\ 
\hline
\end{tabular}}
\vspace{-2mm}
\caption{\textbf{Multi-lingual dataset distribution}}
\label{multi_ling_disribution}
\vspace{-4mm}
\end{table}

\subsection{Bias in Dataset}
\label{app:dataset_bias}
\vspace{-1 ex}
Large-scale models have a profound impact on society, both positive and negative, particularly in applications involving multi-modal tasks. Their performance heavily depends on the datasets they are trained on, and research shows that biases affect certain user groups unfairly or reinforcing discriminatory patterns. Many large-scale models and their training datasets remain inaccessible, with most only available through a restricted input-output interface. While open-source initiatives attempt to replicate model architectures, the lack of publicly available datasets makes it challenging to thoroughly investigate and address potential biases. While bias-free datasets are unattainable \cite{fabbrizzi2022survey}, we ensure our dataset is bias-aware and take active steps toward inclusivity.
\\
\textbf{Selection Bias: }Our primary data sources, international aggregators, naturally emphasize European and American objects, leading to a \textbf{selection bias}, further amplified by the lack of digitization in smaller museums. However, our dataset includes 5M+ objects from other continents, and we are collaborating with local museums to diversify underrepresented cultures. 
\\
\textbf{Temporal bias: }Given the aggregators’ extensive curation, our collection spans a vast historical timeline, from ancient artifacts to modern art, covering statues, paintings, vessels, fossils, corals, war depictions, weapons, manuscripts, textiles, coins, ceramics, scientific instruments, and more. 
\\
\textbf{Language bias: }To mitigate language bias, we include 15M samples across 37 languages beyond English as part of our multilingual dataset with ongoing expansions.
\\
\textbf{Framing Bias: }We also acknowledge framing bias, as models are trained on front-view images as per standard digitization practices, yet multi-angle experiments confirm model robustness to different image perspectives.
\\
\textbf{Bias due to model architecture: } The CLIP model itself introduces biases that are challenging to fully assess, as its training data is not publicly available. Since the vision encoders of both BLIP and LLaVA models rely on CLIP embeddings, they inherit these biases as well. By providing an open large-scale image-text dataset, we enable greater transparency and facilitate the auditing of contrastive image-text models like CLIP.
\\
\textbf{Handling bias in dataset: }
To facilitate a thorough investigation of dataset biases, we will release MUSEUM-65 along with tools designed for large-scale data exploration using precomputed image embeddings \cite{beaumont-2022-clip-retrieval}. These tools will allow researchers to retrieve images based on text or image queries, enabling a systematic study of how biases manifest within the dataset. By examining patterns in object representation, cultural distribution, and framing biases, researchers can gain insights into potential disparities and their implications. In addition to aiding bias detection, these tools will support the development of automated methods for dataset curation, helping mitigate safety concerns associated with large-scale data.
\\
To ensure accessibility, we will publicly release the source code and essential routines, allowing users to build their own versions of these tools for customized dataset exploration. Furthermore, we plan to introduce a public portal where researchers can report undetected biases and model behaviors based on their findings, contributing to ongoing improvements in dataset transparency, fairness, and inclusivity. Through these efforts, we aim to establish MUSEUM-65 as a robust real-world large-scale dataset that not only supports cultural heritage research but also advances responsible data curation practices.
\\
For applications requiring a minimally biased dataset, debiasing techniques such as model-agnostic training or specialized model architectures can also be applied as needed \cite{yuan2021language, ouyang2021suppressing, gu2024beyond} .
\\
\subsection{Safety and Ethical considerations}
Our dataset provides a foundation for training multimodal models that can enhance cultural accessibility, support educational tools, and enable virtual heritage exploration while promoting multilingual data and fostering cross-cultural appreciation by making global artifacts easily comparable.
Upon randomly inspecting images and text from our dataset, we found that museums, as reputable institutions, carefully curate their collections to address potential controversies such as historical disputes, religious issues, privacy concerns, and racial biases. This curation also extends to inappropriate content tagging, ensuring the safety and quality of the dataset. This curation also extends to inappropriate content tagging, ensuring dataset safety and quality. While bias in the place of origin remains an ethical consideration, we aim to mitigate it through collaborations and diversification, with the hope that broader digitization efforts by museums will further enhance dataset diversity
\\
In the current form, we consider this dataset a research artefact and strongly advocate \textbf{ academic use only} and advise careful investigation of downstream model biases. 
\\
This dataset serves as a foundation rather than a final solution for building more balanced and carefully curated datasets for model training. We believe that this process should be open and transparent, involving the broader research community to ensure responsible data development. By introducing MUSEUM-65, a large-scale dataset with diverse image-text pairs and annotations, we provide a resource that can aid in identifying biases, refining data selection, and creating safer, more representative subsets for various applications. We encourage researchers to contribute to this ongoing effort, fostering collaboration toward more ethical and inclusive dataset curation.

\vspace{-2mm}
\section{Experimental details} 
\label{app:exp_details}
\vspace{-1 ex}
This section includes detailed information on the parameters used for fine-tuning, LLaVA and BLIP. It also covers a proposed ablation for LLaVA, experimental configurations, and tuning strategies applied during fine-tuning, providing insights into the optimization process and training evolution of these models. The code can be found: \href{https://github.com/insait-institute/Museum-65}{MUSEUM-65}
\vspace{-1mm}
\subsection{Implementation details}
\vspace{-1 ex}
\textbf{BLIP}
The BLIP model we used is \textit{blip\_vqa}.
During fine-tuning we follow the same protocol as \cite{li2022blip}, having a learning rate 2e-5, a cosine annealing learning rate and the AdamW optimizer \cite{loshchilov2017decoupled}, 
with weight decay 0.05.
We used a batch size of 4x8x16= 512.\\
\textbf{LLaVA}
During fine-tuning we follow the same protocol as \cite{liu2023visualinstructiontuning}, having learning rate 1e-3 and a cosine annealing learning rate schedule with a warmup ratio 0.03 and the AdamW optimizer \cite{loshchilov2017decoupled}, 
with weight decay 0.1. We used LORA for fine-tuning as \cite{liu2023visualinstructiontuning}. We used a batch size of 4x8x16= 512.
\vspace{-1mm}
\subsection{Evaluation metrics}
\vspace{-1 ex}
\label{evaluation_app}
We compute several scores to evaluate our fine-tuned methods for a more diverse assessment. We use both uni-gram and n-gram methods, and choose metrics that are intuitive and well known. 
\\
{\bf Setup.}
To ensure accurate and consistent metric calculations, we pre-process the answers by removing special characters, retaining only alphanumeric content before computing the metrics. The overall metric is an average of individual metric scores for each question.
\\
{\bf Precision.} Given the model's prediction and a list of valid answers for a question, the precision is the fraction of words from the model's prediction that appear in at least one of the valid answers. We consider Complete Precision as the percentage of questions for which the precision is 1.0 (the answer completely matches the ground truth) and Partial Precision as the percentage of questions with precision $>$ 0.0 (the answer partially matches the ground truth).
\\
{\bf Recall.} For each valid answer, the recall is the fraction of words from the answer that are included in model's prediction.
For each question, the recall is averaged among all valid answers. Again, we consider Complete Recall as the percentage of questions for which the recall is 1.0 and Partial Recall as the percentage of questions with recall $>$ 0.0.
\\
{\bf BLEU scores.}
We compute the BLEU score to address matching word pairs accurately. The BLEU score is the fraction of word n-grams from the model's prediction that appear in at least one of the valid answers, modified by a brevity penalty that penalizes short responses that only match a few words. We translate the score to give values between 0 and 100. We compute individual scores for BLEU 1-gram and BLEU 2-gram (referred as BLEU1 and BLEU2) and we average the scores among all the instances.
\vspace{-1mm}
\subsection{Finetuning}
\vspace{-0.5 ex}
\paragraph{Why BLIP and  LLaVA?}
BLIP excels at aligning images with descriptive text, generating accurate captions which contribute to its question answering capabilities, making it a good first choice for VQA. However, BLIP relies on a relatively small pre-trained text encoder/decoder (BERT-base with 110 million parameters), which may limit its depth of understanding, especially for more complex or nuanced instructions and queries. Therefore we also chose the LLaVA model, which uses Llama2 7B, an instruction-tuned LLM which is a much more powerful pre-trained language model that understands instructions better than BLIP.
\vspace{-1 mm}
\begin{table}[h]
\centering
\resizebox{0.87\linewidth}{!}{
\begin{tabular}{lccccc}
\hline
epoch                                & 1    & 2     & 3     & 4     & 5     \\ \hline
LLaVA mQ & 57.3 & 59.51 & 60.75 & 60.77 & 60.77 \\
LLaVA 1Q                         & 54.7 & 55.76 & 56.73 & 57.61 & 58.08 \\ \hline
\end{tabular}}
\vspace{-2mm}
\caption{Comparison of two LLaVA fine-tuning methods: LLaVA-1Q, which uses one random question per image per epoch, and LLaVA-mQ, which utilizes all available questions per image each epoch. LLaVA-mQ achieves better results and faster convergence.}
\label{llava_ablation}
\vspace{-4mm}
\end{table}
\begin{table*}[]
\footnotesize
\resizebox{0.97\linewidth}{!}{
\begin{tabular}{lclclclclc}
\toprule
Subject & \#instances                              & Types    & \#instances                    & Material & \#instances                & Place  & \#instances                       & Creator    & \#instances                    \\ \hline
united states & 2096485 & ship& 96423              & oil& 189599          & united kingdom& 8991283   & Robert John Welch& 4128    \\
university& 845965                            & model& 86022    & canvas& 141786       & texas& 1578768            & British school& 3428       \\
american& 716196                              & vessel& 72338            & paper& 86420         & california& 424987        & William Alfred green& 3352 \\
 maps& 641289          & medal& 47672             & wood& 46831          & massachusetts& 350232     & Joseph Hardman& 2323       \\
school& 561988                                & water transport& 46829   & stone& 46084         & new york& 254190          & John Everett& 1798         \\
church& 419043                                & uniform& 37773           & bronze& 31201        & washington& 253248        & Henry Moore& 1086          \\
river& 337474                                 & artifact& 21469          & glass& 26390         & los angeles& 248918       & Godfrey Kneller& 876       \\
city& 293439                                  & accessory& 18304         & fiber& 14692         & carolina& 177376          & Alfred James Munnings& 731 \\
family& 252846                                & documentary& 15968       & acrylic& 9528        & michigan& 65104           & Joshua Reynolds& 676       \\
company& 213910                               & component& 4463 & steel& 5572 & milwaukee& 54816 & Peter Lely& 629
\\
\bottomrule
\end{tabular}}
\vspace{-1mm}
\caption{Detailed list of the most common values across different categories, \textit{subject, types, material, place, creator} (left), along with the number of instances that correspond to them (right).}
\label{tab:analysis}
\end{table*}
\vspace{-3 mm}
\paragraph{LLaVA ablation}
During fine-tuning we wanted to observe the
impact of using all the questions available for an image and we observed an improvement during evaluation for that model. As it was very time consuming (each epoch being 3 times longer), and as LLaVA already being time expensive, we continued the rest of the experiments with the version that chooses one random question for each image in every epoch. (LLaVA 1Q). See \cref{llava_ablation}.
\vspace{-1mm}
\subsection{Training Evolution}
\begin{figure}
    \centering
    \includegraphics[clip, trim=0cm 0cm 0cm 0cm, width=.45\textwidth]{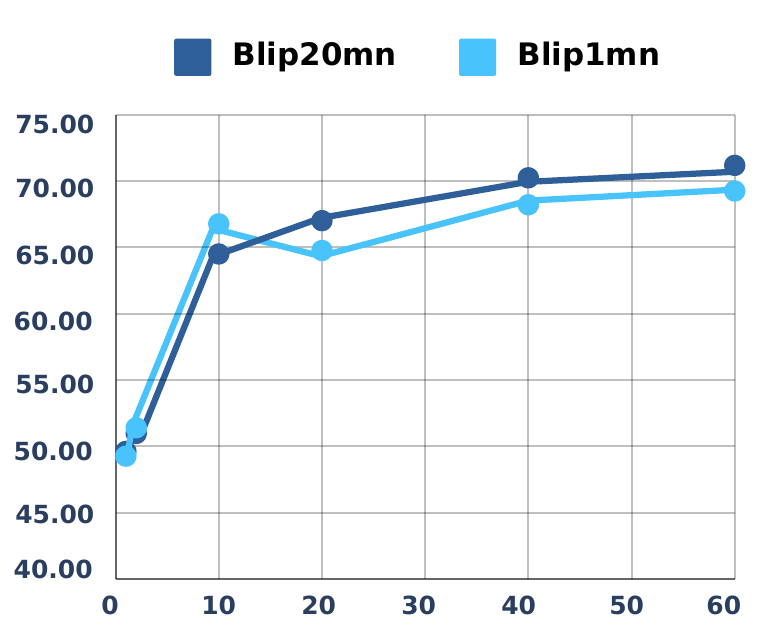}
    \caption{Comparison between BLIP20mn-1ep ad BLIP1mn-20ep across multiple epochs during fine-tuning, maintaining the same number of steps. We observe BLIP20mn-1ep having better results than BLIP1mn-20ep. }
    \label{fig:blip20vs1}
    \vspace{-3mm}
\end{figure}

\vspace{-1 ex}
We present the performance of BLIP across different epochs, highlighting its progression during training. It compares the outcomes of various BLIP and LLaVA fine-tuning approaches, see \cref{tab:epochs}. We also show a comparison between BLIP1mn and BLIP20mn when having the same amounts of steps, meaning BLIP1mn is trained for 20 epochs while BLIP20mn is trained for 1 epoch (BLIP1mn-20ep and BLIP20mn-1ep), see \cref{fig:blip20vs1}. We observe that BLIP20mn-1ep is having better results than BLIP1mn-20ep highlighting that the amount of data matters. 
\begin{table}[]
\centering
\small
\begin{tabular}{lccccc}
\hline
model $\backslash$ epoch                              & 1    & 2 & 3 & 4 & 5 \\ \hline
LLaVA1mn-5ep &54.7&55.76&56.73&57.61&58.08\\
BLIP1mn-5ep                                  &49.24&51.2&56.34&55.54&56.67\\
BLIP10mn-5ep                                 &64.05&66.67&68.49&69.02&69.23\\
BLIP20mn-5ep        &67.03&69&70.23&71.17&71.51 \\
\bottomrule
\end{tabular}
\caption{Comparison of multiple models over 5 epochs, highlighting their performance progression. The results show that LLaVA achieves significantly better outcomes much earlier in training compared to other models.}
\label{tab:epochs}
\vspace{-2mm}
\end{table}


\section{Additional Results}
This section includes supplementary findings, expanding the primary results presented in the main study, more detailed evaluations of the experiments and graphics comparing the performance of multiple models, for a deeper understanding of their strengths and weaknesses. It also provides insights into the questions created specifically for these analyses.
\subsection{Benchmark Comparison}
\label{app:benchmark_comparison}
\vspace{-1 ex}
In order to show the superior utility of our dataset compared to other existing in literature, we also fine-tuned BLIP on MUZE ({MUZE-BLIP}). Our model (BLIP-20m-5e) outperforms MUZE-BLIP significantly in all settings (see \cref{tab:dataset_comparison}). Thanks to the large scale nature of our dataset, model trained on it achieves the best results on MUZE in a zero-shot manner.

\begin{table}[h]
\vspace{-1.8ex}
\centering
\footnotesize
\resizebox{0.97\linewidth}{!}{
\begin{tabular}{llccccc}
\toprule
Model & Test dataset & \makecell{Partial\\Prec.} & \makecell{Complete\\Prec.} & \makecell{Partial\\Recall} & \makecell{Complete\\Recall} & BLEU1 \\
\midrule
\vspace{0.2mm}
MUZE-BLIP      & MUZE & {67.84} & 51.19 & 67.84 & 51.19 & 27.97 \\
Ours-BLIP       & MUZE & \textbf{79.77} & \textbf{66.69} & \textbf{79.77} & \textbf{66.69} & \textbf{40.58} \\
\hline
\vspace{0.5mm}
MUZE-BLIP  & Ours & {36.38} & 25.93  & 36.38 & 25.93 & 18.7  \\
Ours-BLIP      & Ours & \textbf{71.51} & \textbf{60.58}  & \textbf{71.51} & \textbf{33.95}  & \textbf{48.9}  \\
\bottomrule
\end{tabular}
}
\vspace{-2ex}
\caption{BLIP on \textbf{MUZE vs. MUSEUM-65} datasets.}
\label{tab:dataset_comparison}
\vspace{-5.5mm}
\end{table}
\subsection{General VQA}
\vspace{-1 ex}
Following the General Visual Question Answering (VQA) settings, we present a comprehensive table comparing all BLIP and LLaVA models fine-tuned on our dataset evaluated across multiple metrics, see \cref{tab:generalvqa2}. We observe that in general, the fine-tuned models have much better results than the original models. The results show that LLaVA achieves the best performance among the models.
\begin{table*}[]
\centering
\begin{tabular}{lrrrrrrrr}
\toprule
 & \multicolumn{1}{l}{\begin{tabular}[c]{@{}l@{}}partial \\ prec.\end{tabular}} & \multicolumn{1}{l}{\begin{tabular}[c]{@{}l@{}}complete\\ prec.\end{tabular}} & \multicolumn{1}{l}{\begin{tabular}[c]{@{}l@{}}partial \\ recall\end{tabular}} & \multicolumn{1}{l}{\begin{tabular}[c]{@{}l@{}}complete\\ recall\end{tabular}} & \multicolumn{1}{l}{BLEU1} & \multicolumn{1}{l}{BLEU2} & \multicolumn{1}{l}{BLEU3} & \multicolumn{1}{l}{BLEU4} \\ 
\midrule
BLIP         & 9.1                            & 4.65                            & 9.1                         & 6.27                         & 5.76                       & 0.13                       & 0                          & 0                          \\
BLIP1mn-5ep                         & 56.67                 & 43.5                            & 56.67                        & 21.82                        & 34.57                      & 14.01                      & 3.56                       & 2.45                       \\
BLIP1mn-20ep                        & 64.75                           & 53.65                           & 64.74                        & 29.6                         & 43.01                      & 22.37                      & 5.27                       & 3.67                       \\
BLIP1mn-60ep & 69.24                           & 56.97                           & 69.24                        & 31.48                        & 46.08                      & 24.51                      & 6.32                       & 4.37                       \\
BLIP10mn-5ep                        & 69.23                           & 58.18                           & 69.23                        & 32.8                         & 47.16                      & 25.85                      & 6.34                       & 4.38                       \\
BLIP20mn-1ep                        & 67                              & 55.89                           & 67                           & 31.18                        & 45.02                      & 23.91                      & 5.5                        & 3.84                       \\
BLIP20mn-5ep                        & 71.51                           & 60.58                           & 71.51                        & 33.95                        & 48.9                       & 27.22                      & 7.27                       & 5.13                       \\
LLaVA                               & 23                              & 3.97                            & 23                           & 4.07                         & 5.03                       & 0.28                       & 0.1                        & 0.04                       \\
LLaVA1mn-1ep                        & 73.12                           & 56.28                           & 73.18                        & 55.1                         & 50.12                      & 30.74                      & 10.36                      & 6.98                       \\
LLaVA1mn-5ep                        & \textbf{76.27}                  & \textbf{60.04}                  & \textbf{76.31}               & \textbf{59.14}               & \textbf{53.45}             & \textbf{33.5}              & \textbf{12.56}             & \textbf{8.64}              \\
LLaVA20mn-1ep                       & \textbf{81.25}                  & \textbf{63.96}                  & \textbf{81.26}               & \textbf{63.21}               & \textbf{57.06}             & \textbf{36.38}             & \textbf{14.84}             & \textbf{10.38}   \\
\bottomrule
\end{tabular}
\caption{\textbf{General VQA results}. Comparison of all the fine-tuned models and their no fine-tune version on precision and recall. We observe the models fine-tuned with 20mn dataset are obtaining the best results, while \textbf{LLaVA20mn-1ep is the best}, having 80\% of the object with partial precision and 64\% with complete precision. Also the \textbf{LLaVA models seem to have much better results for recall than the BLIP ones}, being similar with the precision results, showing that the prediction of LLaVA models are more often containing or contained in the ground truth.}
\label{tab:generalvqa2}
\end{table*}

\subsection{Category-wise VQA}
\begin{figure}
    \centering
    \includegraphics[clip, trim=0cm 0cm 0cm 0cm, width=0.32\textwidth]{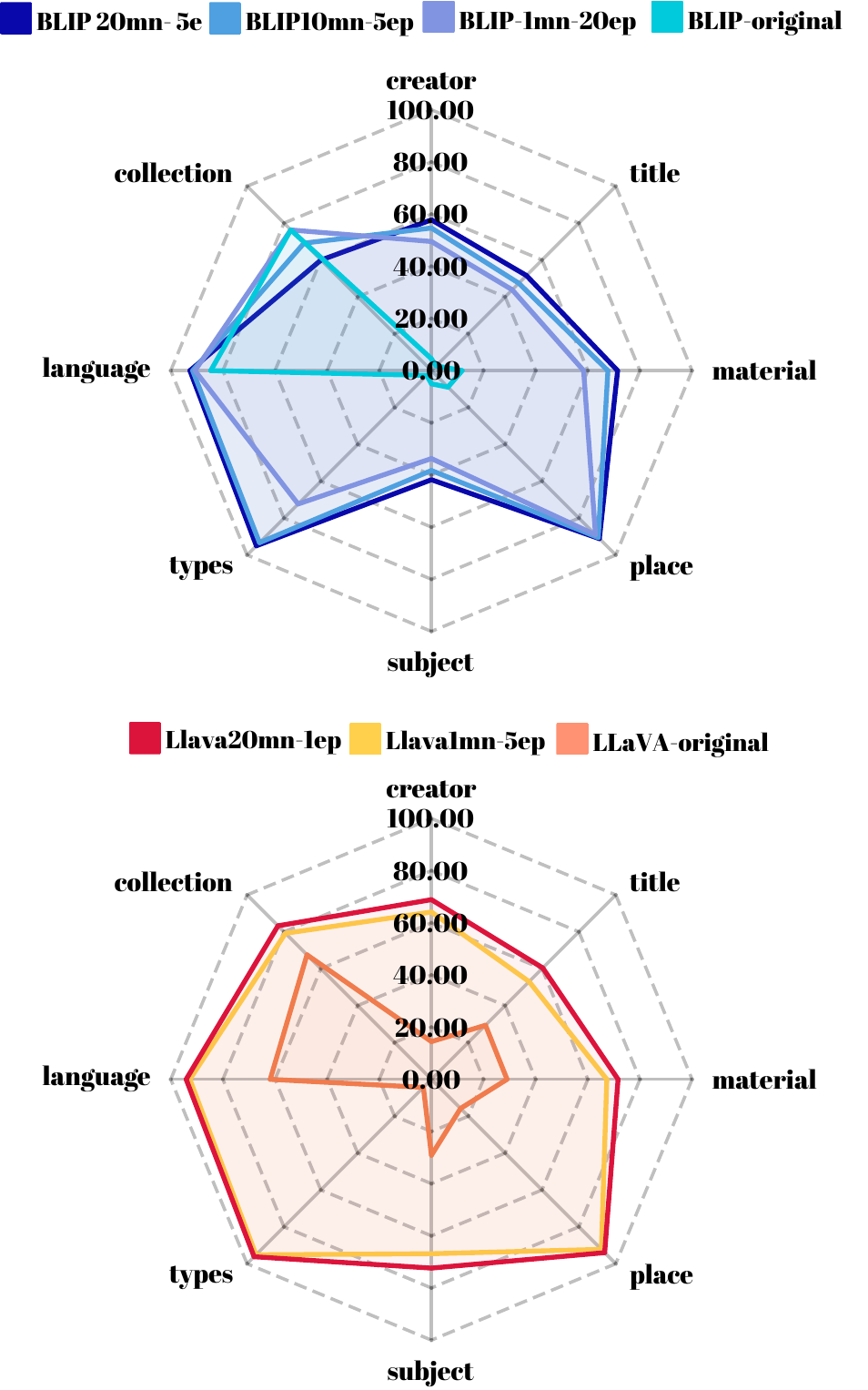}
    \caption{\textbf{VQA category-wise results.} On left be compared all BLIP models  and in right all LLaVA models. The \textbf{fine-tuned models do better on all categories}. The original ones only perform well for \textit{language} and \textit{collection}, as they have easier, common knowledge answers (for \textit{collection}, the results are also influenced by the reduced number of instances that have questions about this). \textbf{LLaVA20mn obtained the best results} among all models, showing significant improvement for \textit{subjects, collection, creator} and \textit{title}, surpassing fine-tuned BLIP.}
    \label{fig:category_wise0}
    \vspace{-5mm}
\end{figure}

\vspace{-1 ex}
For category-wise Visual Question Answering (VQA), we present the results of multiple BLIP and LLaVA models compared with each other across categories such as \textit{subject, title, creator, material} and more (see \cref{fig:category_wise0}). The results demonstrate improved performance of the fine-tuned models in each category. Moreover, the LLaVA fine-tuned models are having better results than BLIP ones on \textit{subject, title, creator, collection, language} and \textit{type}.

\subsection{MultiAngle VQA}
\vspace{-1 ex}
Following the MultiAngle VQA setting, we presents the table comparing multiple models on both original images and images from different viewpoints with extended metrics, helping to evaluate model performance across varying perspectives, offering deeper insights into their robustness. See \cref{tab:multiangle2}
\begin{table*}[]
\small
\centering
\begin{tabular}{lrrrrrrrr}
\toprule
                                  & \multicolumn{1}{l}{\begin{tabular}[c]{@{}l@{}}partial \\ prec.\end{tabular}} & \multicolumn{1}{l}{\begin{tabular}[c]{@{}l@{}}complete\\ prec.\end{tabular}} & \multicolumn{1}{l}{\begin{tabular}[c]{@{}l@{}}partial \\ recall\end{tabular}} & \multicolumn{1}{l}{\begin{tabular}[c]{@{}l@{}}complete\\ recall\end{tabular}} & \multicolumn{1}{l}{BLEU1} & \multicolumn{1}{l}{BLEU2} & \multicolumn{1}{l}{BLEU3} & \multicolumn{1}{l}{BLEU4} \\
                                  \midrule
LLaVA20mn-1ep                     & 58.09                                                                        & 46.09                                                                        & 58.12                                                                         & 41.04                                                                         & 42.14                     & 7.19                      & 2.08                      & 0.52                      \\
\multicolumn{1}{r}{changed angle} & 56.14                                                                        & 44.89                                                                        & 56.15                                                                         & 40.01                                                                         & 41.02                     & 6.97                      & 1.97                      & 0.49                      \\
LLaVA no finetune                 & 24.35                                                                        & 0.09                                                                         & 24.35                                                                         & 11.25                                                                         & 1.61                      & 0.01                      & 0                         & 0                         \\
\multicolumn{1}{r}{changed angle} & 23.56                                                                        & 0.02                                                                         & 23.56                                                                         & 10.85                                                                         & 1.54                      & 0.02                      & 0                         & 0                         \\
BLIP20mn-5ep                      & 52.78                                                                        & 42.51                                                                        & 52.78                                                                         & 35.29                                                                         & 38.31                     & 8.01                      & 1.48                      & 0.24                      \\
\multicolumn{1}{r}{changed angle} & 51.75                                                                        & 41.87                                                                        & 51.75                                                                         & 34.59                                                                         & 37.62                     & 7.84                      & 1.48                      & 0.26                      \\
BLIP no finetune                  & 13.82                                                                        & 9.7                                                                          & 13.82                                                                         & 5.22                                                                          & 6.52                      & 0.02                      & 0.01                      & 0                         \\
\multicolumn{1}{r}{changed angle} & 12.86                                                                        & 8.71                                                                         & 12.86                                                                         & 4.72                                                                          & 5.92                      & 0.01                      & 0                         & 0   \\
\bottomrule
\end{tabular}
\caption{\textbf{MultiAngle results}. Comparing fine-tuned LLaVA20mn-1ep and BLIP20mn-5ep along with the no fine-tune models. We observe the alternative angle images results remain close to the original images results across all metrics for all the models which shows \textbf{stability in regard to changing the angle}, even if the difference between the images is visible.}
\label{tab:multiangle2}
\end{table*}

\begin{table*}[]
\footnotesize
\resizebox{0.97\linewidth}{!}{
\begin{tabular}{ccc}
\toprule
Painters                                                   & Countries                                                                          & Europe                                                                                  \\
\midrule
What is the period the artist lived in?                    & \makecell{Which continent is the country of origin \\of this object located in? }               & \makecell{Which oceans border the continent\\ of origin of this object?}                             \\
What is the nationality of the artist?                     & \makecell{Who are the neighbors of the country of origin\\ of this object?}                     & \makecell{What are the major languages spoken\\ in the continent of origin of this object?}          \\
What is the name of the spouse of the artist?              & \makecell{When did the country of origin of this object \\get independence or get established?} & \makecell{What is the largest country by area \\in the continent of origin of this object?}          \\
Who was the mentor of the artist?                          & \makecell{Which part did the country of origin of \\this object support during World War 2?}    & \makecell{What is the smallest country in the \\continent of origin of this object?}                 \\
Who was influenced by the artist?                          & \makecell{What is the main religion of the country of \\origin of this object?}                 & \makecell{What are some major rivers in the continent\\ of origin of this object?}                   \\
What is the capital of the country the artist was born in? & \makecell{What is the form of government in the country \\of origin of this object?}            & \makecell{What is the dominant climate of the continent\\ of origin of this object?}                 \\
What was the political regime when the artist lived?       & \makecell{Who is the president of the country of origin \\of this country?}                     & \makecell{What are the main religions in the continent \\of origin of this object?}                 \\
Who was the king/president in the period the artist lived? & \makecell{What is the capital of the country of \\origin of this object?}                       & \makecell{What are some of the major economic sectors of \\the continent of origin of this object?}
\\
\bottomrule
\end{tabular}}
\caption{The questions used for the Visually Unanswerable Questions VQA task. These questions are derived from the dataset information starting from the painters or the country of origin for some images. We also added questions related to the continent due to the big number of objects located in Europe, that usually do not have  precise location of origin.}
\label{tab:task4Q}
\end{table*}
\begin{table}[]
\centering
\small
\vspace{-2mm}
\begin{tabular}{ll}
\toprule
Countries      & Artists               \\
\midrule
Germany        & Abdourahmane Sakaly    \\
France         & George Victor Du Noyer \\
USA            & Leo Swan               \\
Netherlands    & Shakespeare William    \\
Italy          & Robert John Welch      \\
Ireland        &                        \\
Denmark        &                        \\
Belgium        &                        \\
United Kingdom &                        \\
Europe         &    \\
\bottomrule
\end{tabular}
\vspace{-2mm}
\caption{The lists of the countries and the artists used for the Visually Unaswerable Questions VQA experiment.}
\label{tab:lists}
\vspace{-5mm}
\end{table}
\begin{table}[]
\centering
\footnotesize
\begin{tabular}{lcccccccc}
\toprule 
\makecell{} &
 \makecell{partial \\ prec.} & \makecell{complete\\ prec.} & \makecell{partial \\ recall} & \makecell{complete\\ recall}  \\
\midrule
LLaVA20mn-1ep      & {10.04}                                & {0.8}                                                                 & 10.02                                                                         & 0.17                                                                           \\
LLaVA nofinetune   & {30.11}                                & {0.27}                                                                  & 30.59                                                                         & 0.56                                                                                              \\
BLIP20mn-5ep & 2.37                                                       & 0.27                                                                                                             & 2.41                                                                          & 0                      \\
BLIP nofinetune    & 1.40                                                        & 0.44                                                                                                               & 1.43                                                                          & 0                                                                         \\ \bottomrule                      
\end{tabular}
\caption{\textbf{MultiLanguage results}. (French and German). We observe that LLaVA models have better results than BLIP ones, still LLaVA20mn-1ep is \textbf{slightly forgetting the ability to answer in other languages}, due to its fine-tuning in English. However, on complete precision and BLEU2 the results of LLaVA20mn-1ep are sligthly better than for the no fine-tune version. }
\vspace{-4mm}
\label{tab:multilang2}
\end{table}

\subsection{Visually Unanswerable Questions VQA}
\label{app:vuq_questions}
\vspace{-1 mm}
We created 510 Q\&A pairs for this task, featuring 5 painters and 10 continents. The dataset includes 5 images per painter and 5 images per country, ensuring a diverse and balanced representation of artists and geographic regions. Each image is paired with 5-8 questions depending on the available information for their subject (painter, country). In \cref{tab:lists} we show the countries and artists used during the experiment and in \cref{tab:task4Q} we present the questions associated with them. As many exhibits where coming from Europe, we included Europe among the countries and designed special questions for it.

\subsection{MultiLanguage VQA}
\vspace{-1 ex}
Following the MultiLanguage VQA setting, we present an extended evaluation of model performance on French and German languages. This analysis provides insights into how well the models handle VQA tasks across different linguistic contexts. See \cref{tab:multilang2}.

\vspace{-0.5 mm}
\section{Limitations and society impact}
\vspace{-2.3 mm}
One limitation of the dataset is that it contains an \textbf{unequal representation of objects} from different cultures or regions, which may introduce bias in training models. This imbalance could lead to under-representation of certain cultural artifacts, affecting the model's ability to generalize well across diverse cultural contexts. Additionally, the \textbf{variability in the quality and depth of information} provided by different museums further complicates the dataset. Some museums may offer detailed descriptions for their objects, while others provide minimal or inconsistent metadata, which could impact the performance of image-text pairing models when dealing with incomplete or sparse information.

\section{Examples}
\begin{figure*}
    \centering
    \includegraphics[clip, trim=0cm 0cm 0cm 0cm, width=1\textwidth]{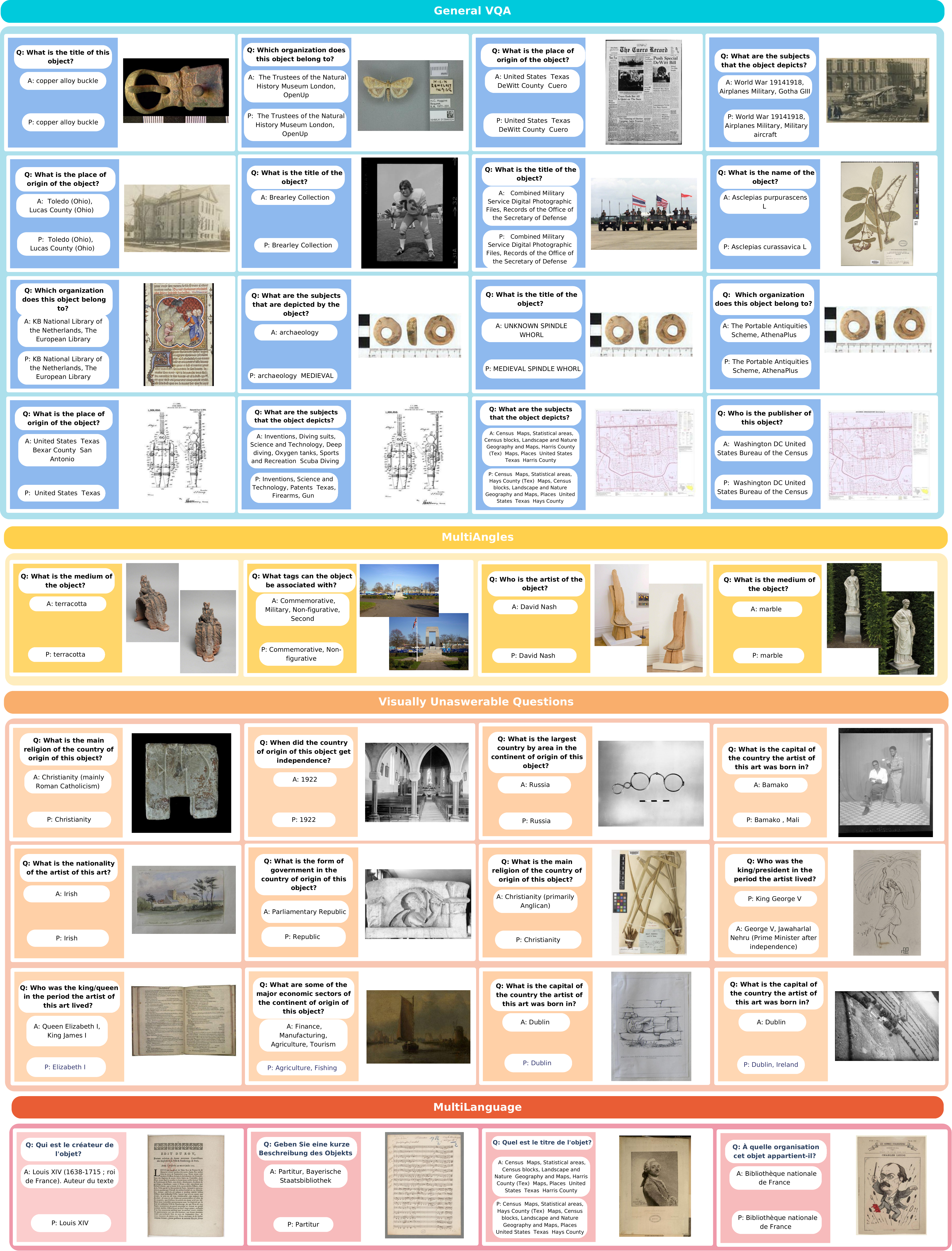}
    \caption{Examples of LLaVA20mn-5ep results for the proposed tasks. The question is denoted with (Q), the answer wit (A) and the prediction with (P).}
    \label{fig:tasks_examples}
\end{figure*}

In \cref{fig:tasks_examples} we show examples of prediction (P) for the best model finetuned with our dataset, LLaVA20mn-5ep, for the proposed tasks.

\end{document}